\documentclass{article}

\usepackage{natbib}
\usepackage{listings} % 在导言区加入这个包

\usepackage{multirow}
\usepackage{graphicx}
\usepackage{placeins} % Import the placeins package to use \FloatBarrier
\usepackage[table,xcdraw]{xcolor}
\usepackage{booktabs}
\usepackage{subcaption} % Include this in your preamble for subtable support
\usepackage[preprint]{neurips_2023}
\usepackage[utf8]{inputenc} % allow utf-8 input
\usepackage[T1]{fontenc}    % use 8-bit T1 fonts
\usepackage{url}            % simple URL typesetting
\usepackage{booktabs}       % professional-quality tables
\usepackage{amsfonts}       % blackboard math symbols
\usepackage{nicefrac}       % compact symbols for 1/2, etc.
\usepackage{microtype}      % microtypography
\usepackage{colortbl}
\usepackage[table]{xcolor}
\definecolor{darkgreen}{HTML}{3cb44b}
\definecolor{darkred}{HTML}{ff7256}
\definecolor{xdcolor}{HTML}{19b5af}
\usepackage{multirow}
\usepackage[hidelinks]{hyperref}
\usepackage{graphicx}
\usepackage{xspace}
\usepackage{geometry}
\usepackage{changepage}
\usepackage{arydshln}
\usepackage{setspace}
\usepackage{multicol}  % 用于双列布局
\newcommand{\llm}{Xmodel-1.5\xspace}

\title{\llm: An 1B-scale Multilingual LLM}
\author{
Wang Qun
\hspace{0.8em}Liu Yang
\hspace{0.8em}Lin Qingquan
\hspace{0.8em}Jiang Ling \\ \\
XiaoduoAI \\
\texttt{\{wangqun,liuyangfoam,linqingquan\}@xiaoduotech.com}
}
\date{}

\begin{document}

\maketitle

\begin{abstract}

We introduce \textbf{\llm}, a 1-billion-parameter multilingual large language model pretrained on 2 trillion tokens, designed for balanced performance and scalability. Unlike most large models that use the BPE tokenizer, \llm employs a custom unigram tokenizer with 65,280 tokens, optimizing both efficiency and accuracy. The model delivers competitive results across multiple languages, including Thai, Arabic, French, Chinese, and English, outperforming Alibaba’s PolyLM-1.7B on respective evaluation datasets. \llm excels in benchmarks like mMMLU and PIQA, and achieves state-of-the-art results in Thai. To support low-resource language research, we release \textbf{Xdata\_Thai}, a Thai-specific evaluation dataset featuring unique linguistic challenges such as gendered particles and idioms. While the model demonstrates strong performance, there is still room for improvement in handling culturally specific nuances. We hope this work contributes to advancements in multilingual AI research. Models and code are publicly available at \url{https://github.com/XiaoduoAILab/XmodelLM-1.5}.

\end{abstract}

\section{Introduction}

The rapid globalization of communication has created an urgent need for advanced multilingual natural language processing (NLP) models that can bridge linguistic divides across regions. Traditional NLP models often struggle with less-represented languages, limiting their effectiveness in global applications. As cross-cultural interactions increase, there is a growing demand for AI systems that can understand and generate multiple languages with high accuracy and relevance.

In response to this challenge, Xiaoduo Technology's AI Lab has developed a 1-billion-parameter multilingual large model. Our model excels not only in widely spoken languages such as Chinese and English, but also in languages like Thai, Arabic, and French, demonstrating top-tier performance among models of similar scale. This work addresses the critical need for more inclusive AI systems capable of serving a wider array of linguistic and cultural contexts.

Beyond the model itself, we also contribute to the research community by open-sourcing a Thai evaluation dataset. This dataset, consisting of hundreds of questions annotated by students from Chulalongkorn University’s School of Integrated Innovation, offers a valuable resource for future research in Thai language processing. These efforts highlight our commitment to advancing multilingual AI and improving the tools available for global communication and research. A detailed description of the dataset creation process is provided in Appendix~\ref{sec:appendix_evaluation_set}.

\section{Related Work on Multilingual Large Language Models}
Multilingual large language models (LLMs) have gained significant attention in recent years, addressing the challenges of natural language processing across diverse languages. These models aim to generalize well across both high-resource and low-resource languages, offering a pathway for improved cross-lingual understanding and generation. Several notable models have contributed to this area, including \textbf{XLM-R}, \textbf{mT5}, and \textbf{PolyLM}, which provide important benchmarks for multilingual AI development.

\textbf{XLM-R (XLM-RoBERTa)} \citep{conneau2020unsupervisedcrosslingualrepresentationlearning} is a widely-used pre-trained language model supporting over 100 languages. With parameter sizes ranging from 270M to 3.5B, XLM-R has set a high standard for natural language understanding tasks such as classification and question answering. Its robust generalization to low-resource languages has been a key factor in its success. The 1B parameter variant of XLM-R serves as an important benchmark for comparing models of similar scale, as it balances strong performance with computational efficiency.

\textbf{mT5 (Multilingual T5)} \citep{xue2021mt5massivelymultilingualpretrained}, developed by Google, is designed for both understanding and generation tasks across more than 100 languages. Ranging from 300M to 13B parameters, mT5 has shown impressive results, particularly in low-resource settings. The 1B parameter version is effective at handling complex cross-lingual tasks like machine translation and question answering, making it a suitable model for comparing performance on languages such as Thai, Arabic, and French.

\textbf{PolyLM} \citep{wei2023polylmopensourcepolyglot}, developed by Alibaba DAMO Academy, is a more recent open-source multilingual model available in two sizes: 1.7B and 13B parameters. PolyLM incorporates bilingual data and utilizes a curriculum learning strategy to progressively introduce more non-English data during training. This approach improves its performance on lower-resource languages such as Thai and Indonesian. PolyLM has shown strong results across a wide range of multilingual tasks, often outperforming models like LLaMA and BLOOM on non-English languages.

\section{Pretraining}
This chapter details the pretraining process of \llm. We begin by introducing the sources and composition of our corpus, followed by an explanation of our preprocessing methods. Next, we describe the construction of our customized tokenizer. Finally, we outline the model architecture and training parameter configurations.

 \subsection{Training Data}

% \textbf{Data sourcing}:We built upon the chinese and english data of the first generation of the Xmodel \citep{wang2024xmodellmtechnicalreport} with the following details: In the process of constructing the training corpus and allocating weights, our primary objective is to ensure the quality and diversity of the training data. Our original dataset primarily consists of aggregated training data from other LLMs, such as Redpajama \citep{together2023redpajama}, subsets of the Pile \citep{gao2020pile} and StarCoder \citep{li2023starcoder}. To address deficiencies in the distribution of book and mathematical data within the training data distribution, we have also incorporated FanFics\footnote{https://huggingface.co/datasets/marianna13/fanfics} and OpenWebMath \citep{paster2023openwebmath}. Additionally, we have added the Chinese data source PTD \citep{wang2024telechat} and WanJuan \citep{he2023wanjuan} to imbue our model with a certain level of proficiency in Chinese.

\begin{figure}[ht]
\centering
\includegraphics[width=\linewidth]{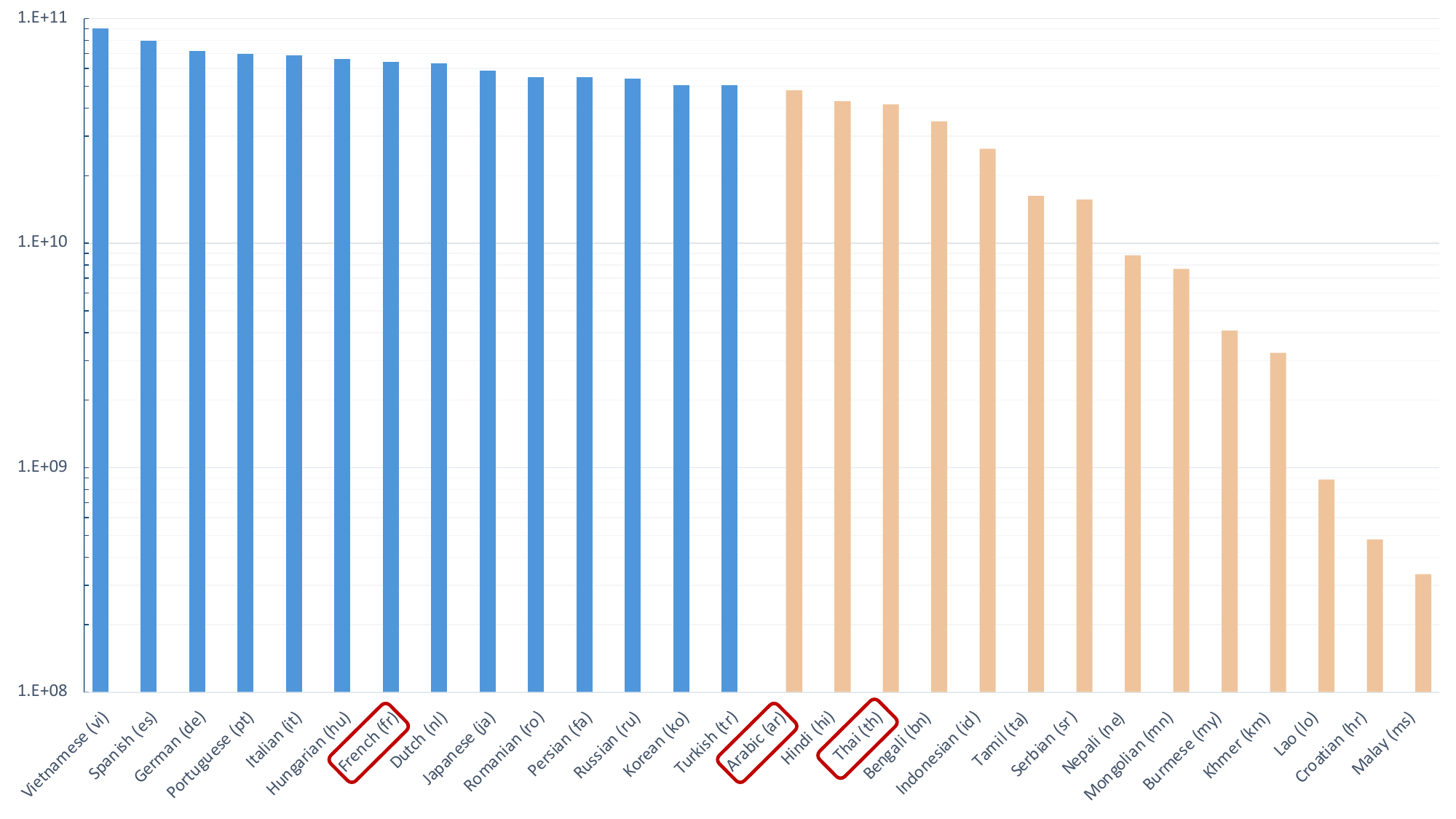}
\caption{Total multilingual data tokens during the pretraining phase sourced from MultiWiki and CulturaX.}
\label{fig:multilang_token_chart}
\end{figure}

\textbf{Data Sourcing}: Building upon the Chinese and English data from the first generation of Xmodel \citep{wang2024xmodellmtechnicalreport}, we focused on enhancing the model’s performance in low-resource languages. To achieve this, we expanded the dataset by incorporating diverse multilingual data, specifically sourced from Multilang Wiki and CulturaX \citep{nguyen2023culturax}, as shown in Figure ~\ref{fig:multilang_token_chart}

The Wiki data was directly downloaded in 30 languages and preprocessed \citep{cyberzhg2018wikidumpreader} to remove unnecessary markers such as URLs. Similarly, CulturaX \citep{nguyen2023culturax} provided data in 27 languages, with an emphasis on oversampling low-resource languages such as Mongolian (mn), Burmese (my), Nepali (ne), Khmer (km), Serbian (sr), and Tamil (ta), despite their smaller data volumes. To ensure data quality, we applied SimHash-based deduplication to the Wiki data, while leaving the CulturaX \citep{nguyen2023culturax} data unchanged.

To further improve the model’s proficiency in Chinese, we incorporated data from PTD \citep{wang2024telechat} and WanJuan \citep{he2023wanjuan}, similar to the previous Xmodel. However, during the annealing phase, we transitioned the WanJuan dataset to a curated version that focuses on e-commerce domain data.
Additionally, during the annealing phase, we introduced high-quality instructional data\citep{phatthiyaphaibun-etal-2023-pythainlp,wangchanthaiinstruct,phatthiyaphaibun_2024_10783421,lowphansirikul2020scb,kobkrit_viriyayudhakorn_2021_4539916,vajirayana_filtered_tlc} , collected with feedback from Thai colleagues. The proportions of this instructional data are provided in Table~\ref{tab: decay_instruct_data_info}.

\begin{figure}[ht]
\centering
\begin{minipage}[t]{0.49\linewidth}
    \raggedright
    \includegraphics[width=\linewidth]{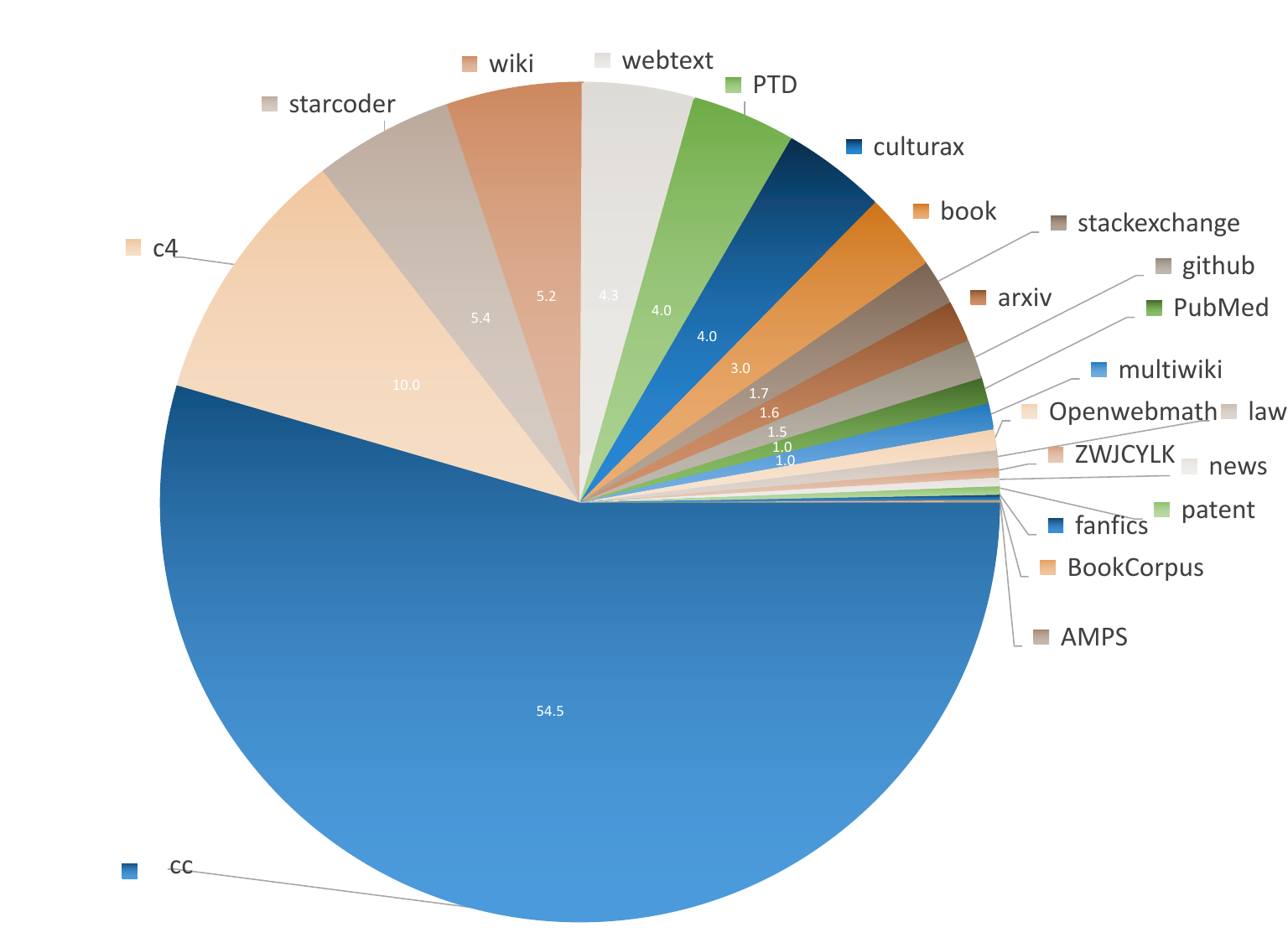}
    \caption{Data distribution during pretraining between 44,000 and 190,000 steps.}
    \label{fig:decay_data1}
\end{minipage}%
\hfill  % 让两个 minipage 之间留有合适的空间
% \hspace{0pt}  % 去除两张图片之间的间距
% \hspace{-0.25\linewidth} % 使用负间距减少总宽度
\begin{minipage}[t]{0.49\linewidth}
    \raggedright
    \includegraphics[width=\linewidth]{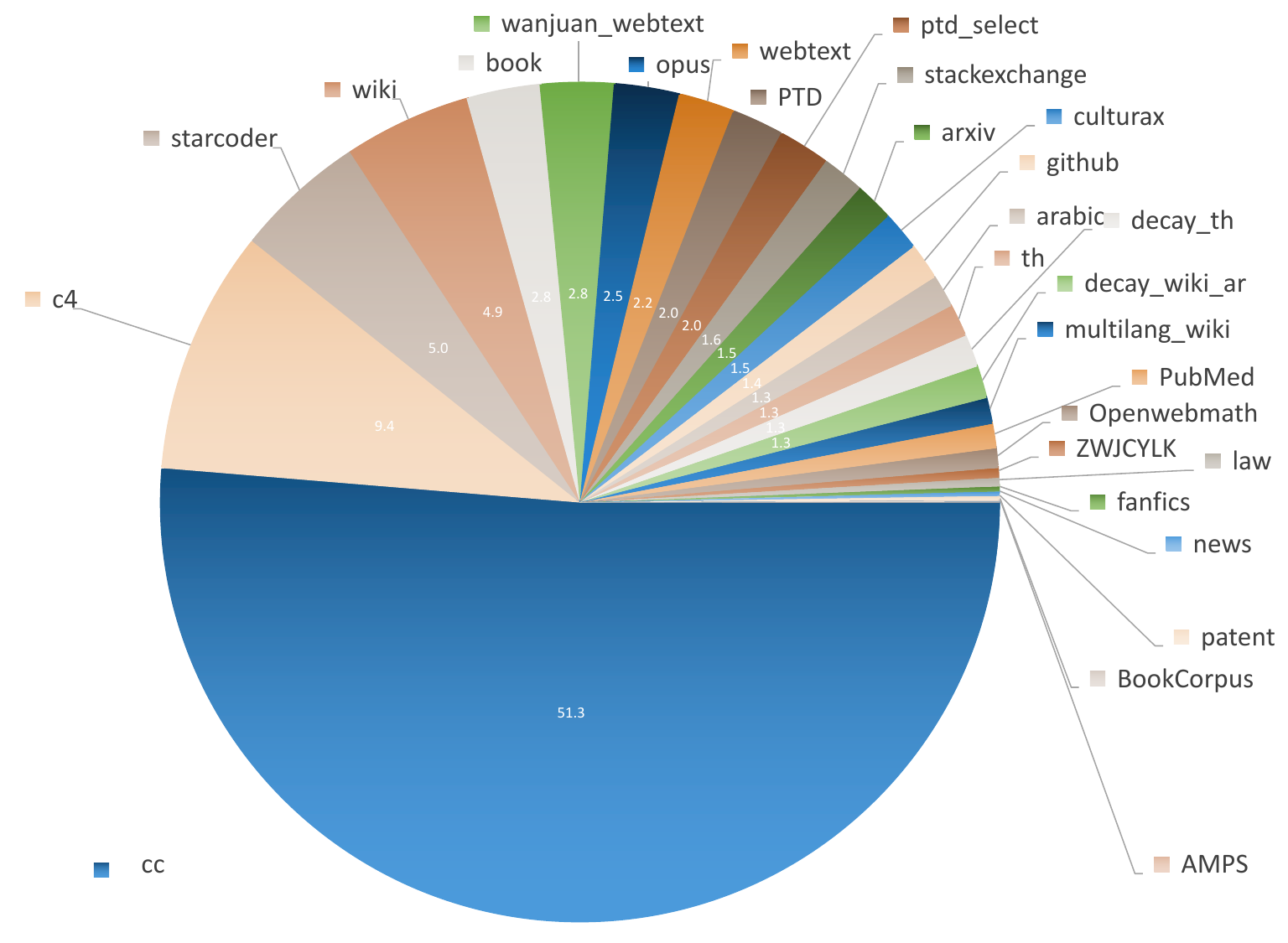}
    \caption{Data distribution during the decay phase.}
    \label{fig:decay_data2}
\end{minipage}
\end{figure}

% \textbf{Data processing}: We are committed to ensure the quality of the data and reducing its redundancy. We first employ heuristic methods such as paragraph length and punctuation ratio for initial filtering. Subsequently, we utilize a 5-gram Kneser-Ney model based on KenLM Library \citep{heafield-2011-kenlm} to compute text perplexity for further quality filtering. In the next stage, we employe a locality-sensitive hashing method based on SimHash to deduplicate the multiligual wiki data. To balance deduplication quality and efficiency, we implement a bucketing strategy on the entire dataset, enabling the deduplication process to scale efficiently across large datasets. Finally, we tokenize all datasets using our custom-trained tokenizer, and designed different sampling weights for the datasets based on their characteristics, as shown in Table~\ref{tab: data_info}.

\begin{table}[ht]
  \centering
  \setlength{\tabcolsep}{2pt}
  \begin{tabular}{@{}llrl@{}}
    \noalign{\hrule height 1.2pt}
    Data Format & Dataset Name & Token Count & Category \\ \midrule

    \multirow{10}{*}{instruct} & klongklon & 12,166,780 & Literature \\ 
    & thai\_usembassy\_th2en\_prompt & 2,000,250 & News, Opus \\ 
    & han-instruct-dataset-v4 & 2,711,072 & Wiki QA, Legal, Opus, Web \\ 
    & scb\_mt\_2020\_th2en\_prompt & 277,208,596 & Opus \\ 
    & thai-wiktionary-prompt & 3,989,318 & Wiktionary \\ 
    & prd\_news\_30112023 & 527,628,430 & News \\ 
    & th\_iapp\_wiki\_qa\_squad & 6,373,522 & Wiki QA \\ 
    & thai-wiki-dataset-v3\_processed\_thaiwikibooks & 8,040,228 & Textbook \\ 
    & oasst2\_thai\_top1\_chat\_format & 93,256 & Multi-task \\ 
    & WangchanThaiInstruct\_processed\_dataset\_thai & 29,366,662 & Medical, Finance, Retail, Legal \\ \midrule
    \multirow{13}{*}{pretrain} & vajirayana\_filtered\_tlc\_content & 103,112,882 & News \\ 
    & wiki\_th & 617,827,100 & Wiki Docs \\ 
    & tnhc & 3,023,992 & Arts \\ 
    & thai\_beginner\_content & 140,652 & Opus \\ 
    & combined\_tlc\_poems & 18,955,824 & Literature \\ 
    & thai-oldbooks & 38,930,572 & Literature \\ 
    & thai-it-books & 579,196 & Tech \\ 
    & goethe-website & 92,162 & Culture \\ 
    & thailand-policy-statements & 1,152,378 & Legal \\ 
    & thai-constitution-corpus & 1,551,736 & Legal \\ 
    & tlcv2.0\_oa\_processed\_raw & 12,687,966 & Literature \\ 
    & thai-financial-dataset & 800,000,000 & Financial \\ 
    \noalign{\hrule height 1.2pt}
  \end{tabular}
  \caption{Detailed Composition of Thai Decay Data.}
  \label{tab: decay_instruct_data_info}
\end{table}

\subsection{Tokenizer}
For our multilingual large model, we used a unigram tokenizer \citep{kudo2018subword} trained with SentencePiece \citep{kudo2018sentencepiece}, resulting in a vocabulary size of 65,280 tokens. This size strikes a balance between performance and efficiency for a 1 billion parameter model, enabling effective handling of diverse languages.

We chose the unigram model \citep{kudo2018unigram} over the commonly used byte pair encoding (BPE) method \citep{bostrom2020bpe} because of its greater flexibility in handling rare and low-frequency tokens, as well as its faster training process. The unigram approach allows more adaptable word segmentation, capturing linguistic nuances and morphological variations, especially for low-resource languages with diverse word forms. While BPE is more memory-efficient, it requires much longer training times. For instance, in our experiments, a BPE model with a 128,000-token vocabulary took 60 hours and 300GB of memory for a 51.2GB dataset with 270 million lines, whereas the unigram model completed the same task in under 12 hours, though it required 1TB of memory due to its sampling process.

After several iterations of tokenizer design, we finalized a version with several key improvements:

\textbf{Training Data Size and Distribution}: The \llm tokenizer was trained on a 50GB subset of the \llm pre-training corpus, with additional industry-specific data to prepare the model for commercial applications. The data distribution was 50\% English, 25\% Chinese, 10\% industry-specific, and 15\% low-resource languages. No additional text normalization was applied.

\textbf{Vocabulary Size}: The vocabulary size was increased from 32,000 to 65,280 tokens to improve the model’s ability to represent diverse languages and specialized terms. Numeric data was encoded by splitting numbers into individual digits. Character coverage was set to 0.9999, with rare characters represented by UTF-8 bytes. To address out-of-vocabulary (OOV) issues, we enabled byte fallback mode and set a maximum token length of 16 to better handle Chinese phrases.

\textbf{Whitespace Handling}: Building on techniques from the LLaMA3 and InternLM2 tokenizers, we manually edited the vocabulary to include tokens for multiple consecutive spaces, improving compression rates for code data. Inspired by MAP-NEO \citep{zhang2024mapneohighlycapabletransparent}, we disabled SentencePiece’s default option to remove extra spaces by setting \texttt{--remove\_extra\_whitespaces=false}, which solved a formatting issue observed in Xmodel-LM's code generation \citep{wang2024xmodellmtechnicalreport}. This adjustment has been incorporated into \llm.

These improvements make the \llm tokenizer particularly well-suited for multilingual tasks, including those involving low-resource languages and code generation, by balancing language coverage, representational capacity, and data processing efficiency. A comparison of the \llm tokenizer with other widely used tokenizers is shown in Table~\ref{tab: tokenizer}, where our tokenizer demonstrates impressive compression rates despite its relatively small size.

% to change %
\begin{table}[!ht]
    \centering
    \setlength{\tabcolsep}{20pt}
    \begin{tabular}{lcc}
		\noalign{\hrule height 1.2pt}
		Tokenizer & Vocab Size & Compression Rate $\downarrow$ \\ 
        \midrule
        LLaMA 3 & 128,000 & 0.3823 \\
		LLaMA 2     & 32,000      & 0.7524          \\
		InternLM 2      & 103,168     & 0.4124            \\
		Baichuan 2  & 125,696      & 0.4103           \\
		\llm  & 65,280     & {\bf 0.3800}          \\ 
        \noalign{\hrule height 1.2pt}
	\end{tabular}
 \newline
	\caption{Comparison of vocabulary size and text compression rate of \llm's tokenizer with other models. Lower values indicate better compression.}\label{table.tokenzier}
 \label{tab: tokenizer}
\end{table}

\subsection{Model architecture}

We built upon the architecture of the first generation of the Xmodel \citep{wang2024xmodellmtechnicalreport} with the following details:

\begin{table}[ht]
  \centering
  \setlength{\tabcolsep}{6pt}
  \begin{tabular}{@{}cccccc@{}}
    \toprule
    Hidden size & Intermediate size & Attention heads &  KV heads & Layers & Context Len\\
    \midrule
    2048 & 5632 & 32 & 4 & 24 & 4096\\
    \bottomrule
  \end{tabular}
  \newline
  \caption{Detailed settings of \llm.}
  \label{tab: LLM_setting}
\end{table}

\noindent\textbf{Rotary Positional Embedding.} We integrate rotary positional embeddings (RoPE) \citep{su2023roformer} at each layer of the network. Additionally, to improve the model's long-context understanding capability, we adopted the approach from \citep{xiong2023longcontext} and adjusted the RoPE base from 10,000 to 500,000.

\noindent\textbf{RMSNorm.} To enhance training stability, we utilize the RMSNorm \citep{Zhang2019RMSNorm} function to normalize the input of each transformer sub-layer, without normalizing the output. Linear layers do not incorporate bias, and word embeddings are not tied.

\noindent\textbf{SwiGLU.} We replace the conventional ReLU non-linearity with the SwiGLU \citep{shazeer2020glu} activation function to optimize performance.

\noindent\textbf{Grouped-query attention.} For efficient training and inference, we employ grouped-query attention (GQA) \citep{ainslie2023gqa}, which incorporates 32 attention heads and 4 KV heads.

\subsection{Training}

Training is performed on a single node using 7 out of the 8 available H800 GPUs. To enhance efficiency and maximize Model FLOPS Utilization (MFU), we utilize Distributed Data Parallel (DDP) and FlashAttention-V2.

We apply cumulative gradient updating, setting a mini-batch size of 4 and using 30 gradient accumulation steps per GPU, resulting in a global batch size of 840 with a sequence length of 4096. This configuration produces a total of 3,440,640 tokens per iteration. Training is conducted over 600,000 iterations, yielding a total token count of 2,064,384,000,000.

We optimize using the AdamW optimizer with a peak learning rate of 6e-4. The learning rate linearly ramps up from zero to the peak over the first 2000 updates, then gradually decreases to 2e-5 following a cosine schedule. At 478K iterations, we introduce an exponential decay to further reduce the learning rate. Throughout training, the effective batch size is approximately 3.5 million tokens, with a weight decay rate of 0.1. We also apply gradient clipping with a threshold of 1.0 to control gradient values.

Data allocation evolves throughout training, with the proportion of multilingual data increasing from 5\% to 10\%, with the initial data ratio as illustrated in Figure~\ref{fig:decay_data1} 

\begin{figure}[ht]
\centering
\includegraphics[width=0.8\linewidth]{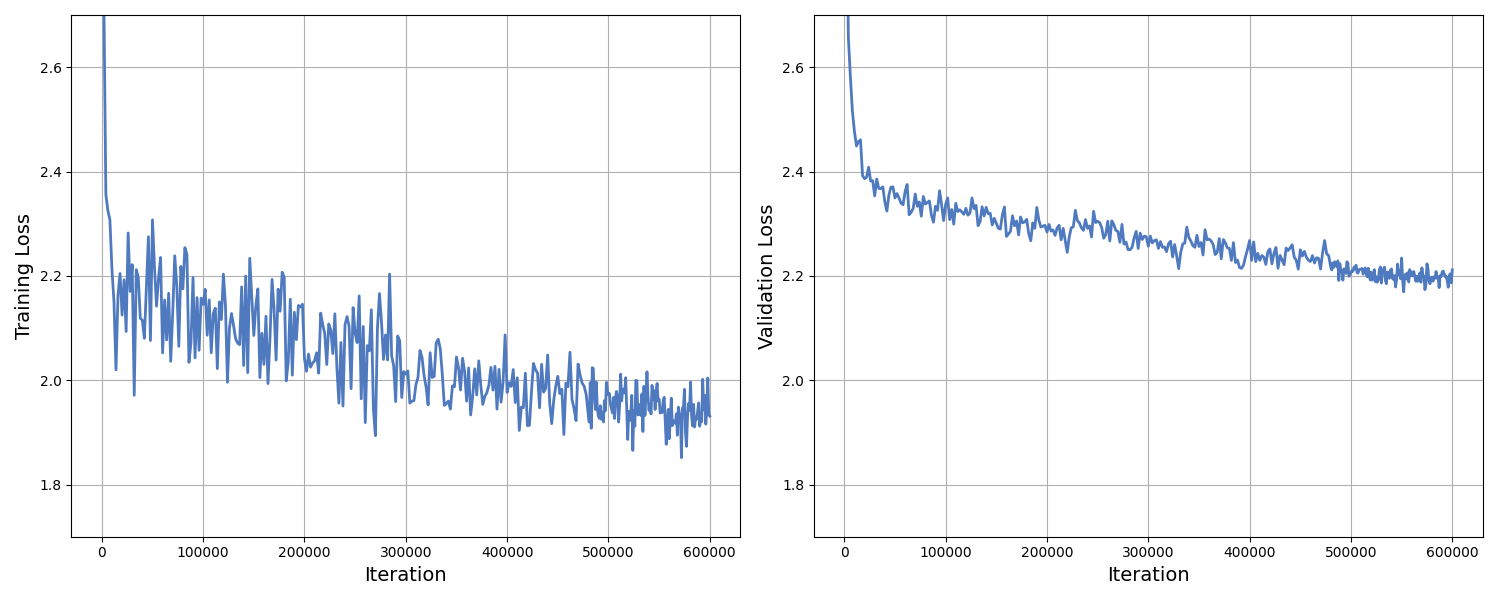}
\caption{The trend of training and validation loss during pretraining.}
\label{fig:loss}
\end{figure}

Refer to Figure~\ref{fig:loss} for training logs, which include trend graphs depicting the progression of training and validation losses as the total token count increases. We use OpenWebText2 \citep{Gokaslan2019OpenWeb}, a dataset excluded from training, as the validation set to measure validation loss.

\section{Post-training}
\label{sec:fine-tuning}

Instruction fine-tuning has become a pivotal approach for improving large pre-trained language models on instruction-based tasks. To enable our model to perform well on e-commerce Retrieval-Augmented Generation (RAG) tasks, we trained an instruction-tuned model using our foundational architecture. Instead of restricting loss computation and backpropagation to just the answer outputs, we utilized the full instruction dataset, resulting in more effective training \citep{shi2024instructiontuninglossinstructions}.

Our fine-tuning setup included a learning rate of 6e-5, weight decay of 0.1, a warmup ratio of 0.03, a context length of 4096 tokens, and a batch size of 120, with a cosine annealing scheduler. We constructed the dataset progressively, starting from simple tasks and advancing to more complex ones, integrating the RAG and RAFT datasets \citep{zhang2024raftadaptinglanguagemodel} toward the end. The dataset comprised Belle (56.04\%) \citep{belle2023exploring}, infinity-instruct-subject (34.25\%), infinity-commonsense (5.97\%), RAG\_mixed (2.97\%), and RAFT\_mixed (0.77\%).

The model's RAG performance in the e-commerce domain achieved a satisfaction rate of 92.47\%, as evaluated by GPT-4o mini on our e-commerce evaluation dataset.

% 指令跟随评测记录：

\section{Evaluation}
\label{sec:evaluation}

\subsection{Evaluating the Pretrained Model}
\noindent\textbf{Baseline Models} To ensure a fair comparison, we selected several popular decoder-only architecture models, each with approximately 1 billion parameters. Specifically, we compare \llm with the following models: OPT \citep{zhang2022opt}, Pythia \citep{biderman2023pythia}, TinyLLaMA \citep{zhang2024tinyllama}, MobileLLaMA \citep{thawakar2024mobillama}, H2O-danube \citep{singer2024h2odanube18b}, InternLM2 \citep{cai2024internlm2} and Qwen2.5 \citep{hui2024qwen25codertechnicalreport}. % Phi-1.5 \citep{li2023textbooks} % StableLM-2 \citep{bellagente2024stable}, MPT \citep{MosaicML2023Introducing} Gemma \citep{gemmateam2024gemma}

\noindent\textbf{Commonsense reasoning tasks} We evaluate our models using the Language Model Evaluation Harness \citep{eval-harness}, which includes tasks such as ARC-Challenge \citep{clark2018think}, ARC-Easy \citep{clark2018think}, Boolq \citep{clark2019boolq}, HellaSwag \citep{zellers2019hellaswag}, OpenBookQA \citep{mihaylov2018suit}, PiQA \citep{Bisk2019PIQARA}, SciQ \citep{welbl2017crowdsourcing}, and Winogrande \citep{Sakaguchi2021WinoGrande}. To ensure fairness and reproducibility, all evaluation metrics were computed in the same environment. Notably, we report raw accuracy metrics, rather than normalized ones. The evaluation results, presented in Table~\ref{tab: commonsense}, show that our model, \llm, outperforms several baseline models, especially surpassing TinyLlama across multiple evaluation metrics.
\begin{table}[ht]
  \centering
  \setlength{\tabcolsep}{3pt}
  \begin{tabular}{lcccccccccc}
    \toprule
    Model & \textbf{ARC-c} & \textbf{ARC-e} & \textbf{Boolq} & \textbf{HS.} & \textbf{OB.} & \textbf{PiQA} & \textbf{SciQ} & \textbf{Wino.} & \textbf{Avg}\\
    \midrule
    \rowcolor{darkgreen!20}
    OPT-1.3B\xspace&23.29&57.03&57.80&41.52&23.20&71.71&84.30&59.59&52.32 \\
    \rowcolor{darkgreen!20}
    Pythia-1.4B\xspace&25.60&57.58&60.34&39.81&20.20&71.06&85.20&56.20&53.38 \\
    % MPT-1.3B\xspace&26.88&60.73&51.8&41.46&23.8&71.38&85.7&11.78&57.3&47.87 \\
    \rowcolor{darkgreen!20}
    TinyLLaMA-3T-1.1B\xspace&27.82&60.31&57.83&44.98&21.80&73.34&88.90&59.12&54.26  \\
    \rowcolor{darkgreen!20}
    MobileLLaMA-1.4B\xspace&26.28&61.32&57.92&42.87&23.60&71.33&87.40&58.25&53.60 \\

    % MobileLLaMA-2.7B-Base\xspace&0.3157&0.6675&0.6367&0.4836&0.272&0.7470&0.920&0.2811&0.6148&0.5487 \\
    % StableLM-2-1.6B\xspace&36.52&66.71&80.09&53.28&26.6&74.86&88.0&7.85&64.25&55.35 \\
    % \rowcolor{darkred!20}
    % H2O-danube-1.8B\xspace&32.94&67.42&65.75&50.85&27.40&75.73&91.50&62.35&59.29 \\
    \rowcolor{darkred!20}
    InternLM2-1.8B\xspace&37.54&70.20&69.48&46.52&24.40&75.57&93.90&65.67&60.41 \\
    \rowcolor{darkred!20}
    Qwen2.5-1.5B\xspace&40.36&74.83&73.27&50.09&31.40&75.95&94.90&63.06&62.98 \\
    % Phi-1.5-1.3B\xspace&44.71&76.14&74.98&47.95&38.6&76.55&93.3&7.96&72.93&59.24 \\
    % Gemma-2B\xspace&40.19&74.24&69.45&52.71&30.20&76.99&94.40&33.23&64.8&59.58 \\
    \midrule
    \rowcolor{xdcolor!20}
    \llm-1B\xspace&28.92&64.31&62.78&45.94&22.20&72.20&89.10&60.62&55.76   \\
    % \rowcolor{xdcolor!20}
    % \llm-1B-Instruct\xspace&32.25&62.37&65.50&46.56&22.20&70.73&90.00&58.96&56.07\\
    \bottomrule
  \end{tabular}
  \newline
  \caption{Performance on commonsense reasoning tasks. Models marked in green perform worse than \llm, while models marked in red perform better than \llm.}
  \label{tab: commonsense}
\end{table}

\begin{figure}[ht]
\centering
\includegraphics[width=\linewidth]{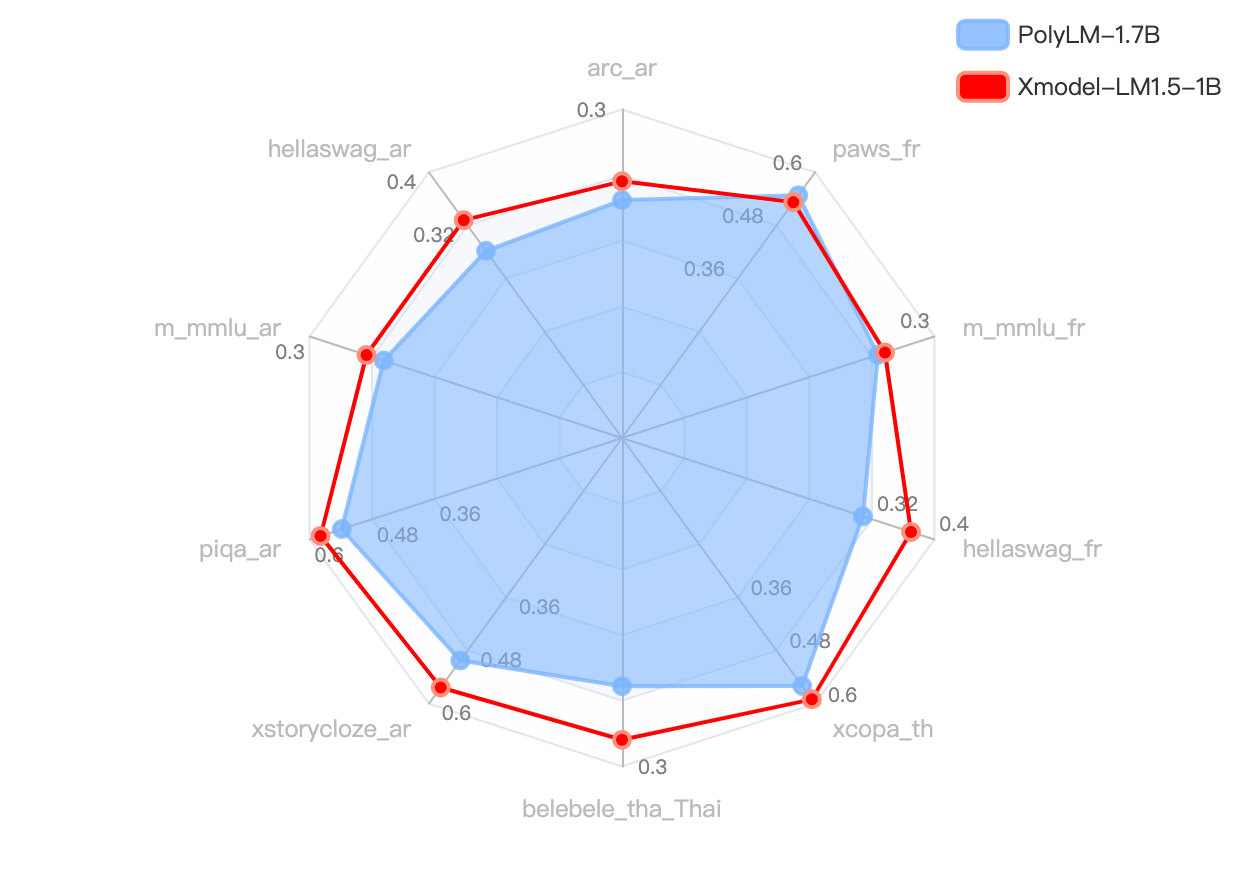}
\caption{Comparison of performance in multilingual tasks between PolyLM 1.7B and \llm 1B}
\label{fig: multilang_radar}
\end{figure}

\textbf{Multilingual ability} In addition to evaluating the model's proficiency in English, we also assessed its multilingual capabilities. Specifically, our evaluation included the following tasks:
\begin{itemize}
%\item C-Eval \citep{huang2023ceval}, this is a comprehensive Chinese evaluation suite for foundation models. It consists of 13948 multi-choice questions spanning 52 diverse disciplines and four difficulty levels.
%\item CMMLU \citep{li2023cmmlu}, this is a comprehensive evaluation benchmark specifically designed to evaluate the knowledge and reasoning abilities of LLMs within the context of Chinese language and culture. CMMLU covers a wide range of subjects, comprising 67 topics that span from elementary to advanced professional levels.
\item ARC \citep{Clark2018ThinkYH}: This dataset consists of 7,787 science exam questions from various sources, including questions provided by a research partner affiliated with AI2. We used the Chinese-translated version of this dataset for our evaluation.
\item XCOPA \citep{ponti2020xcopa}: Designed to test how well machine learning models transfer commonsense reasoning across different languages, XCOPA is a translated and reannotated version of the English COPA \citep{Gordon2011ChoiceOP} and includes 11 languages from different language families and regions worldwide. 

\item {PIQA\_AR} \citep{almazrouei-etal-2023-alghafa}: This is the Arabic version of the PIQA dataset, designed to evaluate physical commonsense reasoning in models, translated by AlGhafa.

\item {Belebele\_tha\_thai} \citep{bandarkar2023belebele}: A Thai subset of the Belebele Benchmark, which assesses multilingual models' reading comprehension through multiple-choice questions based on FLORES-200 passages.

\item{mMMLU}\citep{hendrycks2021measuringmassivemultitasklanguage}: This benchmark evaluates models across 57 tasks, including math, history, and law, highlighting knowledge gaps, particularly in areas like morality and law. 

\item{mHellaswag}\citep{hendrycks2021measuringmassivemultitasklanguage}: A machine-translated version of the HellaSwag \citep{zellers2019hellaswag} dataset, which includes multiple-choice questions to test commonsense reasoning. 

\end{itemize}
The evaluation results are shown in Figure~\ref{fig: multilang_radar}, with further details on the model's performance evolution provided in the case study section~\ref{sec:case_study_evolutions}.

% The evaluation results are presented in Table~\ref{tab: chinese}, all models are evaluated in a zero-shot setting on these tasks. We observed that by adding 15\% Chinese tokens, our model gained a certain degree of understanding and generation capabilities in Chinese, surpassing some existing models, but still weaker compared to InternLM2 and Qwen1.5.

% \begin{table}[ht]
%   \centering
%   \setlength{\tabcolsep}{15pt}
%   \begin{tabular}{lcccc}
%     \toprule
%     Model & \textbf{ARC-zh} & \textbf{XCOPA-zh} & \textbf{XNLI-zh}& \textbf{Avg}\\
%     \midrule
%     % Gemma-2B\xspace   &29.06&61.20&35.10&41.79 \\
%     \rowcolor{darkgreen!20}
%     OPT-1.3B\xspace   &18.80&53.00&33.45&35.08 \\
%     \rowcolor{darkgreen!20}
%     Pythia-1.4B\xspace   &21.03&52.60&34.06&35.90 \\
%     \rowcolor{darkgreen!20}
%     MobileLLaMA-1.4B\xspace   &20.26&52.80&33.82&35.63 \\
%     \rowcolor{darkgreen!20}
%     TinyLLaMA-3T-1.1B\xspace   &21.37&56.80&33.25&37.14 \\
%     \rowcolor{darkgreen!20}
%     H2O-danube-1.8B\xspace  &21.79&55.60&34.74&37.38 \\
%     \rowcolor{darkred!20}
%     InternLM2-1.8B\xspace   &27.69&66.80&34.58&43.00 \\
%     \rowcolor{darkred!20}
%     Qwen1.5-1.8B\xspace &32.14&66.00&39.28&45.81 \\
%     \midrule
%     \rowcolor{xdcolor!20}
%     \llm-1.1B\xspace    &26.24&60.60&36.02&40.95 \\
%     \bottomrule
%   \end{tabular}
%   \newline
%   \caption{Performance on Chinese tasks. Models marked in green perform worse than \llm, while models marked in red perform better than \llm.}
%   \label{tab: chinese}
% \end{table}

\subsection{Evaluation of the Instruction Model}
To evaluate the performance of our instruction model, we conducted a series of standard assessments, including ifeval \citep{zhou2023ifeval}, which tests various aspects of language understanding and instruction-following abilities, as well as MT-Bench \citep{Bai_2024}, a fine-grained benchmark designed to evaluate large language models in multi-turn dialogues. The results, summarized in Table~\ref{tab:instructions}, highlight the model’s strengths and weaknesses in following instructions and managing dialogue complexity.

\begin{table}[ht]
  \centering
  \setlength{\tabcolsep}{20pt}
  \begin{tabular}{lccc}
    \toprule
    Model & \textbf{IFEval} & \textbf{MT-Bench} \\
    \midrule
    
    TinyLlama-1.1B-Chat-v1.0\xspace&5.96&3.46 \\
    Qwen2.5-1.5B-Instruct\xspace&42.5&N/A \\
    StableLM-2-zephyr-1.6B\xspace&32.79&5.42 \\
    H2O-Danube-1.8B-Chat\xspace&15.16&5.52 \\
    InternLM2-Chat-1.8B\xspace&23.87&4.94 \\
    
    Gemma-2B\xspace&20.38&5.19 \\
    Phi-2-2.7B\xspace&27.39&4.29 \\
    PolyLM-Chat-13B \xspace & 16.27&N/A \\
    \midrule
    \llm-Instruct-1B\xspace&3.7&5.06 \\
    \bottomrule
  \end{tabular}
  \newline
  \caption{Performance on instruction following and multi-turns chat tasks. }
  \label{tab:instructions}
\end{table}

\section{Case Study}
% \subsection{Evaluation Insights from Chulalongkorn Collaboration}
% We developed a user-based evaluation and annotation interface for students at Chulalongkorn University, as shown in Figure~\ref{fig:website}. Through this interface, annotators can perform ad hoc tests and rate model outputs. Feedback, including suggested corrections, is stored in the backend database. A brief overview of both good and poor cases is provided in Appendix~\ref{sec:appendix}, which also discusses challenges such as the model’s handling of Thai slang, gender differentiation, and formal vs. informal tone distinctions. These issues can lead to responses that sound unnatural.

\subsection{Evaluation Insights from Chulalongkorn Collaboration}
\label{sec:evaluation_sights}
\begin{figure}[ht]
\centering
\includegraphics[width=0.8\linewidth]{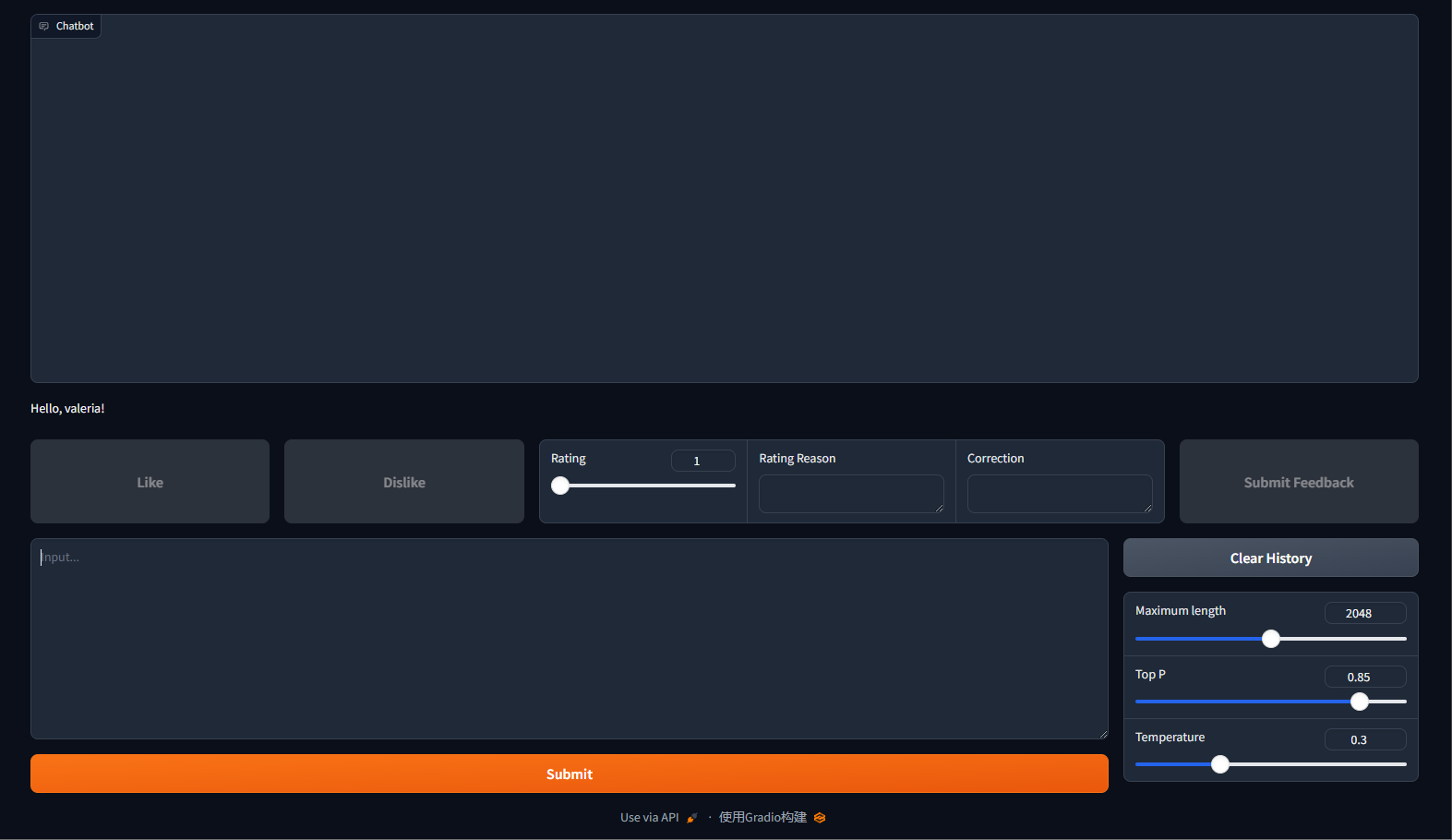}
\caption{Annotation interface for students at Chulalongkorn University}
\label{fig:website}
\end{figure}

We collaborated with Chulalongkorn University to develop a user-based evaluation and annotation interface, as shown in Figure~\ref{fig:website}. This interface enabled students to perform ad hoc testing and rate model outputs, with feedback and suggested corrections stored in a backend database.

Through this collaboration, several key observations were made:
\begin{itemize}
\item The model performed effectively in structured tasks such as e-commerce Q\&A, where responses were described as concise and clear by native speakers (Figure~\ref{fig:efficient_and_concise_answers}).
\item However, challenges were identified in handling Thai-specific linguistic nuances:
\begin{itemize}
    \item \textbf{Gendered Language}: The model struggled to differentiate gendered particles, an issue also observed in other state-of-the-art models (Figure~\ref{fig:sexual_species_bad_case}).
    \item \textbf{Time and Numerical Expressions}: Unique ways of expressing time and numerical data in Thai often resulted in translation errors and misrepresentations (Figure~\ref{fig:time_bad_case}).
\end{itemize}
\end{itemize}

These findings highlight areas for further refinement, particularly in gender differentiation, politeness, and handling culturally specific expressions. A detailed analysis, including examples of good and poor cases, is provided in Appendix~\ref{sec:appendix}.

% \subsection{Cases in Human Evaluation}
% We deployed a user-based model evaluation and annotation interface for students at Chulalongkorn University, as shown in Figure~\ref{fig:website}. Through this interface, annotators can perform ad hoc tests and rate the model's outputs. Their feedback, including reasons and suggested corrections, is directly stored in the backend database. Additionally, present examples of both good and bad cases. The analysis reveals several challenges, such as the model’s limited understanding of Thai slang, its inability to differentiate genders, and its difficulty in distinguishing between formal and informal tones, all of which can result in responses that sound unnatural.

\subsection{Evolution of the Model's Performance}
\label{sec:case_study_evolutions}

\begin{figure}[ht]
\centering
\includegraphics[width=0.8\linewidth]{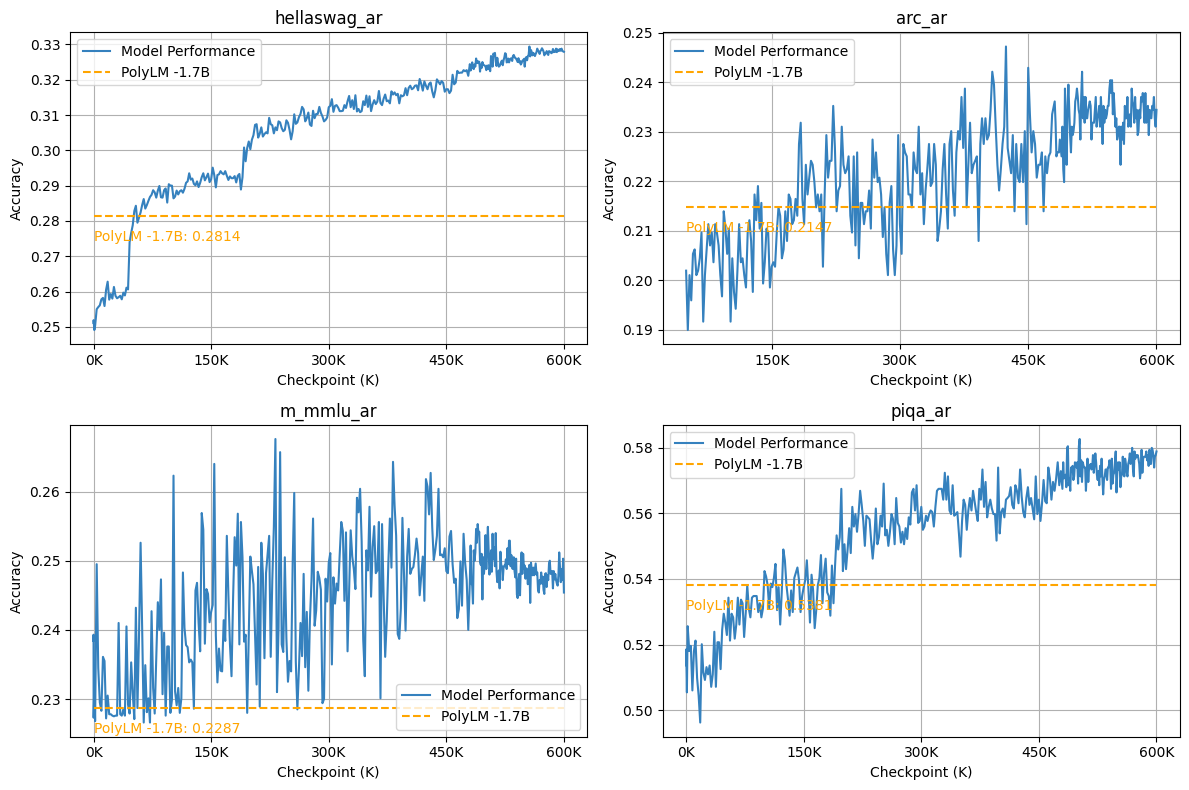}
\caption{Evolution of performance in Arabic bench during pre-training}
\label{fig:ar_evolution}
\end{figure}

\begin{figure}[ht]
\centering
\includegraphics[width=0.8\linewidth]{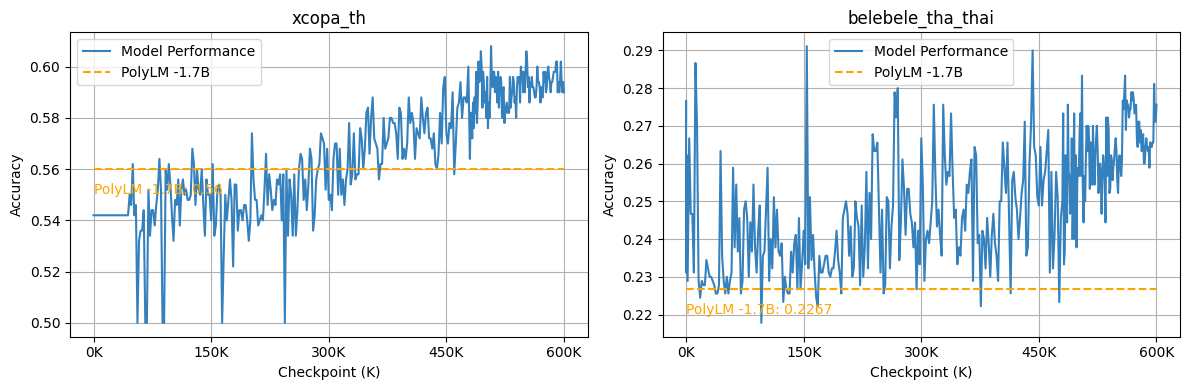}
\caption{Evolution of performance in Thai bench during pre-training}
\label{fig:th_evolution}
\end{figure}

\begin{figure}[ht]
\centering
\includegraphics[width=\linewidth]{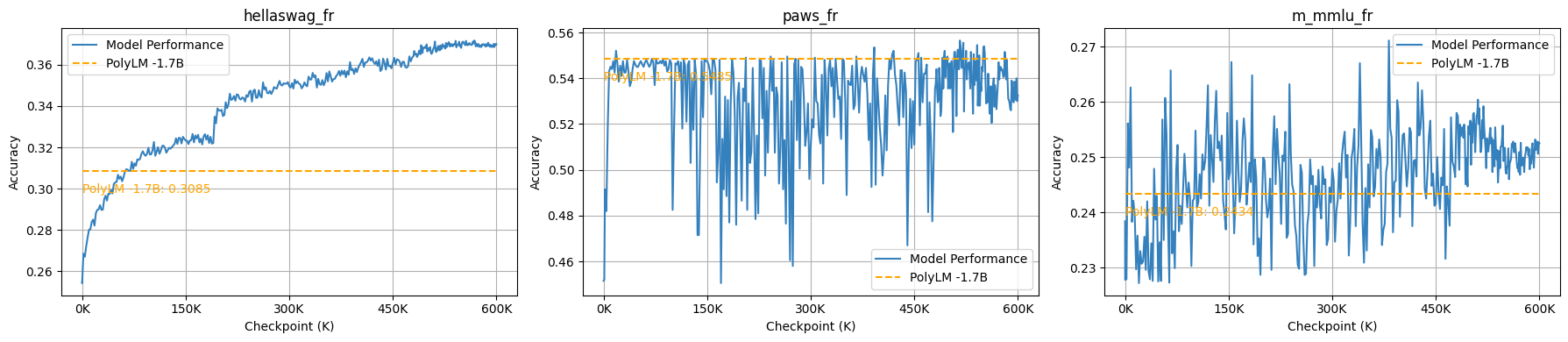}
\caption{Evolution of performance in French bench during pre-training}
\label{fig:fr_evolution}
\end{figure}

We monitored and recorded the model’s performance on the multilingual benchmark throughout the pretraining process, as shown in Figure~\ref{fig:ar_evolution},~\ref{fig:th_evolution},~\ref{fig:fr_evolution}. As training progressed, it became evident that the performance of \llm consistently improved, ultimately surpassing PolyLM-1.7B \citep{wei2023polylmopensourcepolyglot} across multiple tasks.
% Moreover, we observe an approximate linear link between log iteration steps and model metrics gains across most tasks.

% \FloatBarrier % Ensures that all figures are placed before proceeding

\section{Conclusions}

% In summary, our 1 billion parameter multilingual large model marks a valuable exploration in the field of cross-linguistic small language models (SLMs). It demonstrates strong performance across Thai, Arabic, French, and other languages, highlighting its potential to significantly enhance multilingual AI systems and cross-cultural communication. These results not only validate the effectiveness of our approach but also set the stage for further improvements in handling complex linguistic nuances. We are confident that this work will drive future advancements in multilingual AI and play a pivotal role in bridging language barriers.

In summary, our 1-billion-parameter multilingual large language model represents a valuable step forward in the exploration of cross-linguistic small language models (SLMs). It demonstrates strong performance across languages such as Thai, Arabic, and French, showcasing its potential to significantly enhance multilingual AI systems and cross-cultural communication. These results validate the effectiveness of our approach and highlight the model's promise in addressing low-resource language challenges and domain-specific applications, including e-commerce.
While encouraged by these achievements, we recognize opportunities for further refinement, particularly in handling complex linguistic nuances such as gendered expressions, temporal data, and culturally specific idioms. This work provides a foundation for ongoing advancements in multilingual AI, paving the way for more inclusive, accurate, and culturally aligned systems. We are confident that these contributions will play a pivotal role in bridging language barriers and fostering greater understanding across diverse linguistic and cultural landscapes.

\section{Acknowledgments}
We would like to express our sincere gratitude to Dr. Pietro Borsano and students from the School of Integrated Innovation at Chulalongkorn University in Thailand for their invaluable contributions to the Thai large language model collaboration project. Their enthusiasm and dedication, particularly in annotating the Thai evaluation dataset, were instrumental to the success of this work. We also extend our thanks to our colleagues at Xiaoduo Technology, including Xu Zongliang, Shi Yu, Chen Nanxi, and Chen Xinjie, for their efforts in organizing and supporting the project. Their commitment and teamwork were crucial to making this achievement possible. Finally, we would like to thank Qin Shuo for his helpful suggestions on improving the clarity of our writing.

% \FloatBarrier % Ensures that all figures are placed before proceeding

\begin{spacing}{0.9} % 设置行间距稍微紧凑\begin{spacing}{0.9} % 设置行间距稍微紧凑
\footnotesize % 设置较小的字体
\begin{multicols}{2} % 双列布局

\bibliography{xmodel-1.5}

\begin{thebibliography}{52}
\providecommand{\natexlab}[1]{#1}
\providecommand{\url}[1]{\texttt{#1}}
\expandafter\ifx\csname urlstyle\endcsname\relax
  \providecommand{\doi}[1]{doi: #1}\else
  \providecommand{\doi}{doi: \begingroup \urlstyle{rm}\Url}\fi

\bibitem[Ainslie et~al.(2023)Ainslie, Lee-Thorp, de~Jong, Zemlyanskiy, Lebrón, and Sanghai]{ainslie2023gqa}
Joshua Ainslie, James Lee-Thorp, Michiel de~Jong, Yury Zemlyanskiy, Federico Lebrón, and Sumit Sanghai.
\newblock Gqa: Training generalized multi-query transformer models from multi-head checkpoints, 2023.

\bibitem[AIRESEARCH(2023)]{wangchanthaiinstruct}
AIRESEARCH.
\newblock Wangchanthaiinstruct: A thai instruction-following dataset.
\newblock \url{https://huggingface.co/datasets/airesearch/WangchanThaiInstruct}, 2023.
\newblock URL \url{https://huggingface.co/datasets/airesearch/WangchanThaiInstruct}.
\newblock Accessed: 2024-11-15.

\bibitem[Almazrouei et~al.(2023)Almazrouei, Cojocaru, Baldo, Malartic, Alobeidli, Mazzotta, Penedo, Campesan, Farooq, Alhammadi, Launay, and Noune]{almazrouei-etal-2023-alghafa}
Ebtesam Almazrouei, Ruxandra Cojocaru, Michele Baldo, Quentin Malartic, Hamza Alobeidli, Daniele Mazzotta, Guilherme Penedo, Giulia Campesan, Mugariya Farooq, Maitha Alhammadi, Julien Launay, and Badreddine Noune.
\newblock {A}l{G}hafa evaluation benchmark for {A}rabic language models.
\newblock In Hassan Sawaf, Samhaa El-Beltagy, Wajdi Zaghouani, Walid Magdy, Ahmed Abdelali, Nadi Tomeh, Ibrahim Abu~Farha, Nizar Habash, Salam Khalifa, Amr Keleg, Hatem Haddad, Imed Zitouni, Khalil Mrini, and Rawan Almatham, editors, \emph{Proceedings of ArabicNLP 2023}, pages 244--275, Singapore (Hybrid), December 2023. Association for Computational Linguistics.
\newblock \doi{10.18653/v1/2023.arabicnlp-1.21}.
\newblock URL \url{https://aclanthology.org/2023.arabicnlp-1.21}.

\bibitem[Bai et~al.(2024)Bai, Liu, Bu, He, Liu, Zhou, Lin, Su, Ge, Zheng, and Ouyang]{Bai_2024}
Ge~Bai, Jie Liu, Xingyuan Bu, Yancheng He, Jiaheng Liu, Zhanhui Zhou, Zhuoran Lin, Wenbo Su, Tiezheng Ge, Bo~Zheng, and Wanli Ouyang.
\newblock Mt-bench-101: A fine-grained benchmark for evaluating large language models in multi-turn dialogues.
\newblock In \emph{Proceedings of the 62nd Annual Meeting of the Association for Computational Linguistics (Volume 1: Long Papers)}, page 7421–7454. Association for Computational Linguistics, 2024.
\newblock \doi{10.18653/v1/2024.acl-long.401}.
\newblock URL \url{http://dx.doi.org/10.18653/v1/2024.acl-long.401}.

\bibitem[Bandarkar et~al.(2023)Bandarkar, Liang, Muller, Artetxe, Shukla, Husa, Goyal, Krishnan, Zettlemoyer, and Khabsa]{bandarkar2023belebele}
Lucas Bandarkar, Davis Liang, Benjamin Muller, Mikel Artetxe, Satya~Narayan Shukla, Donald Husa, Naman Goyal, Abhinandan Krishnan, Luke Zettlemoyer, and Madian Khabsa.
\newblock The belebele benchmark: a parallel reading comprehension dataset in 122 language variants, 2023.

\bibitem[Biderman et~al.(2023)Biderman, Schoelkopf, Anthony, Bradley, O'Brien, Hallahan, Khan, Purohit, Prashanth, Raff, Skowron, Sutawika, and van~der Wal]{biderman2023pythia}
Stella Biderman, Hailey Schoelkopf, Quentin Anthony, Herbie Bradley, Kyle O'Brien, Eric Hallahan, Mohammad~Aflah Khan, Shivanshu Purohit, USVSN~Sai Prashanth, Edward Raff, Aviya Skowron, Lintang Sutawika, and Oskar van~der Wal.
\newblock Pythia: A suite for analyzing large language models across training and scaling, 2023.

\bibitem[Bisk et~al.(2019)Bisk, Zellers, Bras, Gao, and Choi]{Bisk2019PIQARA}
Yonatan Bisk, Rowan Zellers, Ronan~Le Bras, Jianfeng Gao, and Yejin Choi.
\newblock Piqa: Reasoning about physical commonsense in natural language.
\newblock In \emph{AAAI Conference on Artificial Intelligence}, 2019.
\newblock URL \url{https://api.semanticscholar.org/CorpusID:208290939}.

\bibitem[Bostrom and Durrett(2020)]{bostrom2020bpe}
Kaj Bostrom and Greg Durrett.
\newblock Byte pair encoding is suboptimal for language model pretraining, 2020.
\newblock URL \url{https://arxiv.org/abs/2004.03720}.

\bibitem[Cai et~al.(2024)Cai, Cao, Chen, Chen, Chen, Chen, Chen, Chen, Chen, Chu, Dong, Duan, Fan, Fei, Gao, Ge, Gu, Gu, Gui, Guo, Guo, He, Hu, Huang, Jiang, Jiao, Jin, Lei, Li, Li, Li, Li, Li, Li, Liu, Liu, Hong, Liu, Liu, Liu, Lv, Lv, Lv, Ma, Ma, Ma, Ning, Ouyang, Qiu, Qu, Shang, Shao, Song, Song, Sui, Sun, Sun, Tang, Wang, Wang, Wang, Wang, Wang, Wang, Wang, Wei, Weng, Wu, Xiong, Xu, Xu, Yan, Yan, Yang, Ye, Ying, Yu, Yu, Zang, Zhang, Zhang, Zhang, Zhang, Zhang, Zhang, Zhang, Zhang, Zhang, Zhang, Zhang, Zhao, Zhao, Zhao, Zhou, Zhou, Zhuo, Zou, Qiu, Qiao, and Lin]{cai2024internlm2}
Zheng Cai, Maosong Cao, Haojiong Chen, Kai Chen, Keyu Chen, Xin Chen, Xun Chen, Zehui Chen, Zhi Chen, Pei Chu, Xiaoyi Dong, Haodong Duan, Qi~Fan, Zhaoye Fei, Yang Gao, Jiaye Ge, Chenya Gu, Yuzhe Gu, Tao Gui, Aijia Guo, Qipeng Guo, Conghui He, Yingfan Hu, Ting Huang, Tao Jiang, Penglong Jiao, Zhenjiang Jin, Zhikai Lei, Jiaxing Li, Jingwen Li, Linyang Li, Shuaibin Li, Wei Li, Yining Li, Hongwei Liu, Jiangning Liu, Jiawei Hong, Kaiwen Liu, Kuikun Liu, Xiaoran Liu, Chengqi Lv, Haijun Lv, Kai Lv, Li~Ma, Runyuan Ma, Zerun Ma, Wenchang Ning, Linke Ouyang, Jiantao Qiu, Yuan Qu, Fukai Shang, Yunfan Shao, Demin Song, Zifan Song, Zhihao Sui, Peng Sun, Yu~Sun, Huanze Tang, Bin Wang, Guoteng Wang, Jiaqi Wang, Jiayu Wang, Rui Wang, Yudong Wang, Ziyi Wang, Xingjian Wei, Qizhen Weng, Fan Wu, Yingtong Xiong, Chao Xu, Ruiliang Xu, Hang Yan, Yirong Yan, Xiaogui Yang, Haochen Ye, Huaiyuan Ying, Jia Yu, Jing Yu, Yuhang Zang, Chuyu Zhang, Li~Zhang, Pan Zhang, Peng Zhang, Ruijie Zhang, Shuo Zhang, Songyang Zhang, Wenjian Zhang,
  Wenwei Zhang, Xingcheng Zhang, Xinyue Zhang, Hui Zhao, Qian Zhao, Xiaomeng Zhao, Fengzhe Zhou, Zaida Zhou, Jingming Zhuo, Yicheng Zou, Xipeng Qiu, Yu~Qiao, and Dahua Lin.
\newblock Internlm2 technical report, 2024.

\bibitem[Clark et~al.(2019)Clark, Lee, Chang, Kwiatkowski, Collins, and Toutanova]{clark2019boolq}
Christopher Clark, Kenton Lee, Ming-Wei Chang, Tom Kwiatkowski, Michael Collins, and Kristina Toutanova.
\newblock Boolq: Exploring the surprising difficulty of natural yes/no questions, 2019.

\bibitem[Clark et~al.(2018{\natexlab{a}})Clark, Cowhey, Etzioni, Khot, Sabharwal, Schoenick, and Tafjord]{Clark2018ThinkYH}
Peter Clark, Isaac Cowhey, Oren Etzioni, Tushar Khot, Ashish Sabharwal, Carissa Schoenick, and Oyvind Tafjord.
\newblock Think you have solved question answering? try arc, the ai2 reasoning challenge.
\newblock \emph{ArXiv}, abs/1803.05457, 2018{\natexlab{a}}.

\bibitem[Clark et~al.(2018{\natexlab{b}})Clark, Cowhey, Etzioni, Khot, Sabharwal, Schoenick, and Tafjord]{clark2018think}
Peter Clark, Isaac Cowhey, Oren Etzioni, Tushar Khot, Ashish Sabharwal, Carissa Schoenick, and Oyvind Tafjord.
\newblock Think you have solved question answering? try arc, the ai2 reasoning challenge, 2018{\natexlab{b}}.

\bibitem[Conneau et~al.(2020)Conneau, Khandelwal, Goyal, Chaudhary, Wenzek, Guzmán, Grave, Ott, Zettlemoyer, and Stoyanov]{conneau2020unsupervisedcrosslingualrepresentationlearning}
Alexis Conneau, Kartikay Khandelwal, Naman Goyal, Vishrav Chaudhary, Guillaume Wenzek, Francisco Guzmán, Edouard Grave, Myle Ott, Luke Zettlemoyer, and Veselin Stoyanov.
\newblock Unsupervised cross-lingual representation learning at scale, 2020.
\newblock URL \url{https://arxiv.org/abs/1911.02116}.

\bibitem[CyberZHG(2023)]{cyberzhg2018wikidumpreader}
CyberZHG.
\newblock wiki-dump-reader.
\newblock \url{https://github.com/CyberZHG/wiki-dump-reader}, 2023.
\newblock Accessed: 2024-10-23.

\bibitem[Gao et~al.(2023)Gao, Tow, Abbasi, Biderman, Black, DiPofi, Foster, Golding, Hsu, Le~Noac'h, Li, McDonell, Muennighoff, Ociepa, Phang, Reynolds, Schoelkopf, Skowron, Sutawika, Tang, Thite, Wang, Wang, and Zou]{eval-harness}
Leo Gao, Jonathan Tow, Baber Abbasi, Stella Biderman, Sid Black, Anthony DiPofi, Charles Foster, Laurence Golding, Jeffrey Hsu, Alain Le~Noac'h, Haonan Li, Kyle McDonell, Niklas Muennighoff, Chris Ociepa, Jason Phang, Laria Reynolds, Hailey Schoelkopf, Aviya Skowron, Lintang Sutawika, Eric Tang, Anish Thite, Ben Wang, Kevin Wang, and Andy Zou.
\newblock A framework for few-shot language model evaluation, 12 2023.
\newblock URL \url{https://zenodo.org/records/10256836}.

\bibitem[Gokaslan et~al.(2019)Gokaslan, Cohen, Pavlick, and Tellex]{Gokaslan2019OpenWeb}
Aaron Gokaslan, Vanya Cohen, Ellie Pavlick, and Stefanie Tellex.
\newblock Openwebtext corpus.
\newblock \url{http://Skylion007.github.io/OpenWebTextCorpus}, 2019.

\bibitem[Gordon et~al.(2011)Gordon, Kozareva, and Roemmele]{Gordon2011ChoiceOP}
Andrew~S. Gordon, Zornitsa Kozareva, and Melissa Roemmele.
\newblock Choice of plausible alternatives: An evaluation of commonsense causal reasoning.
\newblock In \emph{AAAI Spring Symposium: Logical Formalizations of Commonsense Reasoning}, 2011.
\newblock URL \url{https://api.semanticscholar.org/CorpusID:434646}.

\bibitem[He et~al.(2023)He, Jin, Xu, Qiu, Wang, Li, Yan, Wang, and Lin]{he2023wanjuan}
Conghui He, Zhenjiang Jin, Chao Xu, Jiantao Qiu, Bin Wang, Wei Li, Hang Yan, Jiaqi Wang, and Dahua Lin.
\newblock Wanjuan: A comprehensive multimodal dataset for advancing english and chinese large models, 2023.

\bibitem[Hendrycks et~al.(2021)Hendrycks, Burns, Basart, Zou, Mazeika, Song, and Steinhardt]{hendrycks2021measuringmassivemultitasklanguage}
Dan Hendrycks, Collin Burns, Steven Basart, Andy Zou, Mantas Mazeika, Dawn Song, and Jacob Steinhardt.
\newblock Measuring massive multitask language understanding, 2021.
\newblock URL \url{https://arxiv.org/abs/2009.03300}.

\bibitem[Hui et~al.(2024)Hui, Yang, Cui, Yang, Liu, Zhang, Liu, Zhang, Yu, Dang, Yang, Men, Huang, Ren, Ren, Zhou, and Lin]{hui2024qwen25codertechnicalreport}
Binyuan Hui, Jian Yang, Zeyu Cui, Jiaxi Yang, Dayiheng Liu, Lei Zhang, Tianyu Liu, Jiajun Zhang, Bowen Yu, Kai Dang, An~Yang, Rui Men, Fei Huang, Xingzhang Ren, Xuancheng Ren, Jingren Zhou, and Junyang Lin.
\newblock Qwen2.5-coder technical report, 2024.
\newblock URL \url{https://arxiv.org/abs/2409.12186}.

\bibitem[Ji et~al.(2023)Ji, Deng, Gong, Peng, Niu, Zhang, Ma, and Li]{belle2023exploring}
Yunjie Ji, Yong Deng, Yan Gong, Yiping Peng, Qiang Niu, Lei Zhang, Baochang Ma, and Xiangang Li.
\newblock Exploring the impact of instruction data scaling on large language models: An empirical study on real-world use cases.
\newblock \emph{arXiv preprint arXiv:2303.14742}, 2023.

\bibitem[Johannes~Welbl(2017)]{SciQ}
Matt~Gardner Johannes~Welbl, Nelson F.~Liu.
\newblock Crowdsourcing multiple choice science questions, 2017.

\bibitem[Kudo(2018{\natexlab{a}})]{kudo2018subword}
Taku Kudo.
\newblock Subword regularization: Improving neural network translation models with multiple subword candidates, 2018{\natexlab{a}}.

\bibitem[Kudo(2018{\natexlab{b}})]{kudo2018unigram}
Taku Kudo.
\newblock Subword regularization: Improving neural network translation models with multiple subword candidates, 2018{\natexlab{b}}.
\newblock URL \url{https://arxiv.org/abs/1804.10959}.

\bibitem[Kudo and Richardson(2018)]{kudo2018sentencepiece}
Taku Kudo and John Richardson.
\newblock Sentencepiece: A simple and language independent subword tokenizer and detokenizer for neural text processing, 2018.

\bibitem[Lowphansirikul et~al.(2020)Lowphansirikul, Polpanumas, Rutherford, and Nutanong]{lowphansirikul2020scb}
Lalita Lowphansirikul, Charin Polpanumas, Attapol~T Rutherford, and Sarana Nutanong.
\newblock scb-mt-en-th-2020: A large english-thai parallel corpus.
\newblock \emph{arXiv preprint arXiv:2007.03541}, 2020.

\bibitem[Mihaylov et~al.(2018)Mihaylov, Clark, Khot, and Sabharwal]{mihaylov2018suit}
Todor Mihaylov, Peter Clark, Tushar Khot, and Ashish Sabharwal.
\newblock Can a suit of armor conduct electricity? a new dataset for open book question answering, 2018.

\bibitem[Nguyen et~al.(2023)Nguyen, Nguyen, Lai, Man, Ngo, Dernoncourt, Rossi, and Nguyen]{nguyen2023culturax}
Thuat Nguyen, Chien~Van Nguyen, Viet~Dac Lai, Hieu Man, Nghia~Trung Ngo, Franck Dernoncourt, Ryan~A. Rossi, and Thien~Huu Nguyen.
\newblock Culturax: A cleaned, enormous, and multilingual dataset for large language models in 167 languages, 2023.

\bibitem[Phatthiyaphaibun(2024)]{phatthiyaphaibun_2024_10783421}
Wannaphong Phatthiyaphaibun.
\newblock Thai tnhc2 books, 2024.
\newblock URL \url{https://doi.org/10.5281/zenodo.10783421}.

\bibitem[Phatthiyaphaibun et~al.(2023)Phatthiyaphaibun, Chaovavanich, Polpanumas, Suriyawongkul, Lowphansirikul, Chormai, Limkonchotiwat, Suntorntip, and Udomcharoenchaikit]{phatthiyaphaibun-etal-2023-pythainlp}
Wannaphong Phatthiyaphaibun, Korakot Chaovavanich, Charin Polpanumas, Arthit Suriyawongkul, Lalita Lowphansirikul, Pattarawat Chormai, Peerat Limkonchotiwat, Thanathip Suntorntip, and Can Udomcharoenchaikit.
\newblock {P}y{T}hai{NLP}: {T}hai natural language processing in {P}ython.
\newblock In Liling Tan, Dmitrijs Milajevs, Geeticka Chauhan, Jeremy Gwinnup, and Elijah Rippeth, editors, \emph{Proceedings of the 3rd Workshop for Natural Language Processing Open Source Software (NLP-OSS 2023)}, pages 25--36, Singapore, Singapore, December 2023. Empirical Methods in Natural Language Processing.
\newblock URL \url{https://aclanthology.org/2023.nlposs-1.4}.

\bibitem[Ponti et~al.(2020)Ponti, Glava\v{s}, Majewska, Liu, Vuli\'{c}, and Korhonen]{ponti2020xcopa}
Edoardo~M. Ponti, Goran Glava\v{s}, Olga Majewska, Qianchu Liu, Ivan Vuli\'{c}, and Anna Korhonen.
\newblock {XCOPA: A} multilingual dataset for causal commonsense reasoning.
\newblock In \emph{Proceedings of the 2020 Conference on Empirical Methods in Natural Language Processing (EMNLP)}, 2020.
\newblock URL \url{https://ducdauge.github.io/files/xcopa.pdf}.

\bibitem[Sakaguchi et~al.(2021)Sakaguchi, Bras, Bhagavatula, and Choi]{Sakaguchi2021WinoGrande}
Keisuke Sakaguchi, Ronan~Le Bras, Chandra Bhagavatula, and Yejin Choi.
\newblock Winogrande: an adversarial winograd schema challenge at scale.
\newblock \emph{Commun. ACM}, 64\penalty0 (9):\penalty0 99–106, aug 2021.
\newblock ISSN 0001-0782.
\newblock \doi{10.1145/3474381}.
\newblock URL \url{https://doi.org/10.1145/3474381}.

\bibitem[Sawatphol(2019)]{vajirayana_filtered_tlc}
Jitkapat Sawatphol.
\newblock Thai literature corpora.
\newblock \\url{https://attapol.github.io/tlc.html}, 2019.

\bibitem[Shazeer(2020)]{shazeer2020glu}
Noam Shazeer.
\newblock Glu variants improve transformer, 2020.

\bibitem[Shi et~al.(2024)Shi, Yang, Wu, Aitchison, Yilmaz, and Lipani]{shi2024instructiontuninglossinstructions}
Zhengyan Shi, Adam~X. Yang, Bin Wu, Laurence Aitchison, Emine Yilmaz, and Aldo Lipani.
\newblock Instruction tuning with loss over instructions, 2024.
\newblock URL \url{https://arxiv.org/abs/2405.14394}.

\bibitem[Singer et~al.(2024)Singer, Pfeiffer, Babakhin, Jeblick, Dhankhar, Fodor, and Ambati]{singer2024h2odanube18b}
Philipp Singer, Pascal Pfeiffer, Yauhen Babakhin, Maximilian Jeblick, Nischay Dhankhar, Gabor Fodor, and Sri~Satish Ambati.
\newblock H2o-danube-1.8b technical report, 2024.

\bibitem[Su et~al.(2023)Su, Lu, Pan, Murtadha, Wen, and Liu]{su2023roformer}
Jianlin Su, Yu~Lu, Shengfeng Pan, Ahmed Murtadha, Bo~Wen, and Yunfeng Liu.
\newblock Roformer: Enhanced transformer with rotary position embedding, 2023.

\bibitem[Thawakar et~al.(2024)Thawakar, Vayani, Khan, Cholakal, Anwer, Felsberg, Baldwin, Xing, and Khan]{thawakar2024mobillama}
Omkar Thawakar, Ashmal Vayani, Salman Khan, Hisham Cholakal, Rao~M. Anwer, Michael Felsberg, Tim Baldwin, Eric~P. Xing, and Fahad~Shahbaz Khan.
\newblock Mobillama: Towards accurate and lightweight fully transparent gpt, 2024.

\bibitem[Viriyayudhakorn and Polpanumas(2021)]{kobkrit_viriyayudhakorn_2021_4539916}
Kobkrit Viriyayudhakorn and Charin Polpanumas.
\newblock iapp\_wiki\_qa\_squad, February 2021.
\newblock URL \url{https://doi.org/10.5281/zenodo.4539916}.

\bibitem[Wang et~al.(2024{\natexlab{a}})Wang, Liu, Yan, Wang, Huang, and Jiang]{wang2024xmodellmtechnicalreport}
Yichuan Wang, Yang Liu, Yu~Yan, Qun Wang, Xucheng Huang, and Ling Jiang.
\newblock Xmodel-lm technical report, 2024{\natexlab{a}}.
\newblock URL \url{https://arxiv.org/abs/2406.02856}.

\bibitem[Wang et~al.(2024{\natexlab{b}})Wang, Liu, Liu, Yao, Huang, He, Li, Li, Che, Zhang, Wang, Wang, Pu, Xu, Fang, Zhao, Zhang, Huang, Lu, Peng, Zheng, Wang, Yang, he, Jiang, Xie, Zhang, Li, Shi, Fu, Zhang, Huang, Xiong, Zhang, Wang, and Song]{wang2024telechat}
Zihan Wang, Xinzhang Liu, Shixuan Liu, Yitong Yao, Yuyao Huang, Zhongjiang He, Xuelong Li, Yongxiang Li, Zhonghao Che, Zhaoxi Zhang, Yan Wang, Xin Wang, Luwen Pu, Huihan Xu, Ruiyu Fang, Yu~Zhao, Jie Zhang, Xiaomeng Huang, Zhilong Lu, Jiaxin Peng, Wenjun Zheng, Shiquan Wang, Bingkai Yang, Xuewei he, Zhuoru Jiang, Qiyi Xie, Yanhan Zhang, Zhongqiu Li, Lingling Shi, Weiwei Fu, Yin Zhang, Zilu Huang, Sishi Xiong, Yuxiang Zhang, Chao Wang, and Shuangyong Song.
\newblock Telechat technical report, 2024{\natexlab{b}}.

\bibitem[Wei et~al.(2023)Wei, Wei, Lin, Li, Zhang, Ren, Li, Wan, Cao, Xie, Hu, Li, Hui, Yu, Liu, Yang, Huang, and Xie]{wei2023polylmopensourcepolyglot}
Xiangpeng Wei, Haoran Wei, Huan Lin, Tianhao Li, Pei Zhang, Xingzhang Ren, Mei Li, Yu~Wan, Zhiwei Cao, Binbin Xie, Tianxiang Hu, Shangjie Li, Binyuan Hui, Bowen Yu, Dayiheng Liu, Baosong Yang, Fei Huang, and Jun Xie.
\newblock Polylm: An open source polyglot large language model, 2023.
\newblock URL \url{https://arxiv.org/abs/2307.06018}.

\bibitem[Welbl et~al.(2017)Welbl, Liu, and Gardner]{welbl2017crowdsourcing}
Johannes Welbl, Nelson~F. Liu, and Matt Gardner.
\newblock Crowdsourcing multiple choice science questions, 2017.

\bibitem[Xiong et~al.(2023)Xiong, Liu, Molybog, Zhang, Bhargava, Hou, Martin, Rungta, Sankararaman, Oguz, Khabsa, Fang, Mehdad, Narang, Malik, Fan, Bhosale, Edunov, Lewis, Wang, and Ma]{xiong2023longcontext}
Wenhan Xiong, Jingyu Liu, Igor Molybog, Hejia Zhang, Prajjwal Bhargava, Rui Hou, Louis Martin, Rashi Rungta, Karthik~Abinav Sankararaman, Barlas Oguz, Madian Khabsa, Han Fang, Yashar Mehdad, Sharan Narang, Kshitiz Malik, Angela Fan, Shruti Bhosale, Sergey Edunov, Mike Lewis, Sinong Wang, and Hao Ma.
\newblock Effective long-context scaling of foundation models, 2023.
\newblock URL \url{https://arxiv.org/abs/2309.16039}.

\bibitem[Xue et~al.(2021)Xue, Constant, Roberts, Kale, Al-Rfou, Siddhant, Barua, and Raffel]{xue2021mt5massivelymultilingualpretrained}
Linting Xue, Noah Constant, Adam Roberts, Mihir Kale, Rami Al-Rfou, Aditya Siddhant, Aditya Barua, and Colin Raffel.
\newblock mt5: A massively multilingual pre-trained text-to-text transformer, 2021.
\newblock URL \url{https://arxiv.org/abs/2010.11934}.

\bibitem[Zellers et~al.(2019)Zellers, Holtzman, Bisk, Farhadi, and Choi]{zellers2019hellaswag}
Rowan Zellers, Ari Holtzman, Yonatan Bisk, Ali Farhadi, and Yejin Choi.
\newblock Hellaswag: Can a machine really finish your sentence?, 2019.

\bibitem[Zhang and Sennrich(2019)]{Zhang2019RMSNorm}
Biao Zhang and Rico Sennrich.
\newblock \emph{Root mean square layer normalization}.
\newblock Curran Associates Inc., Red Hook, NY, USA, 2019.

\bibitem[Zhang et~al.(2024{\natexlab{a}})Zhang, Qu, Liu, Zhang, Lin, Yu, Pan, Cheng, Liu, Lin, Yuan, Zheng, Pang, Du, Liang, Ma, Li, Ma, Lin, Benetos, Yang, Zhou, Ma, Liu, Niu, Wang, Que, Liu, Liu, Guo, Gao, Zhou, Zhang, Zhou, Wang, Bai, Zhang, Zhang, Wang, Yang, Zhao, Zhang, Ouyang, Huang, and Chen]{zhang2024mapneohighlycapabletransparent}
Ge~Zhang, Scott Qu, Jiaheng Liu, Chenchen Zhang, Chenghua Lin, Chou~Leuang Yu, Danny Pan, Esther Cheng, Jie Liu, Qunshu Lin, Raven Yuan, Tuney Zheng, Wei Pang, Xinrun Du, Yiming Liang, Yinghao Ma, Yizhi Li, Ziyang Ma, Bill Lin, Emmanouil Benetos, Huan Yang, Junting Zhou, Kaijing Ma, Minghao Liu, Morry Niu, Noah Wang, Quehry Que, Ruibo Liu, Sine Liu, Shawn Guo, Soren Gao, Wangchunshu Zhou, Xinyue Zhang, Yizhi Zhou, Yubo Wang, Yuelin Bai, Yuhan Zhang, Yuxiang Zhang, Zenith Wang, Zhenzhu Yang, Zijian Zhao, Jiajun Zhang, Wanli Ouyang, Wenhao Huang, and Wenhu Chen.
\newblock Map-neo: Highly capable and transparent bilingual large language model series, 2024{\natexlab{a}}.
\newblock URL \url{https://arxiv.org/abs/2405.19327}.

\bibitem[Zhang et~al.(2024{\natexlab{b}})Zhang, Zeng, Wang, and Lu]{zhang2024tinyllama}
Peiyuan Zhang, Guangtao Zeng, Tianduo Wang, and Wei Lu.
\newblock Tinyllama: An open-source small language model, 2024{\natexlab{b}}.

\bibitem[Zhang et~al.(2022)Zhang, Roller, Goyal, Artetxe, Chen, Chen, Dewan, Diab, Li, Lin, et~al.]{zhang2022opt}
Susan Zhang, Stephen Roller, Naman Goyal, Mikel Artetxe, Moya Chen, Shuohui Chen, Christopher Dewan, Mona Diab, Xian Li, Xi~Victoria Lin, et~al.
\newblock Opt: Open pre-trained transformer language models.
\newblock \emph{arXiv preprint arXiv:2205.01068}, 2022.

\bibitem[Zhang et~al.(2024{\natexlab{c}})Zhang, Patil, Jain, Shen, Zaharia, Stoica, and Gonzalez]{zhang2024raftadaptinglanguagemodel}
Tianjun Zhang, Shishir~G. Patil, Naman Jain, Sheng Shen, Matei Zaharia, Ion Stoica, and Joseph~E. Gonzalez.
\newblock Raft: Adapting language model to domain specific rag, 2024{\natexlab{c}}.
\newblock URL \url{https://arxiv.org/abs/2403.10131}.

\bibitem[Zhou et~al.(2023)Zhou, Lu, Mishra, Brahma, Basu, Luan, Zhou, and Hou]{zhou2023ifeval}
Jeffrey Zhou, Tianjian Lu, Swaroop Mishra, Siddhartha Brahma, Sujoy Basu, Yi~Luan, Denny Zhou, and Le~Hou.
\newblock Instruction-following evaluation for large language models, 2023.
\newblock URL \url{https://arxiv.org/abs/2311.07911}.

\end{thebibliography}

\end{multicols}
\end{spacing}

% \newpage

\section{Appendix}

\subsection{Project Collaboration with Chulalongkorn University}
\label{sec:appendix}
This section presents the results and observations from our project collaboration with Chulalongkorn University.

 % Figures ~\ref{fig:efficient_and_concise_answers}, ~\ref{fig:sexual_species_bad_case},~\ref{fig:time_bad_case} % 
% We collaborated with Chulalongkorn University to deploy an evaluation and annotation system for students, allowing them to test and rate the model’s output. Feedback was directly stored in the backend, helping us identify areas for improvement.

% Figure ~\ref{fig:website} shows the interface used by students. It allowed them to interact with the model and provide feedback.

\begin{figure}[ht]
\centering
\includegraphics[width=0.8\linewidth]{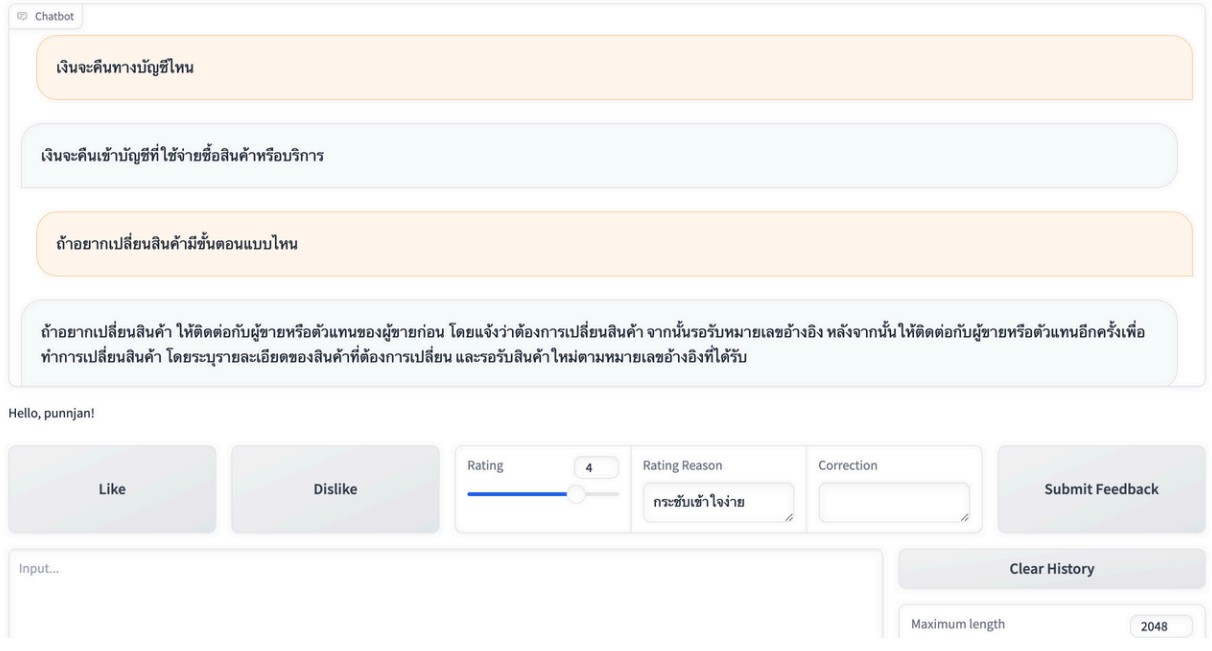}
% \caption{We found that \llm performs effectively in the e-commerce QA context. Feedback from students at Chulalongkorn University described it as "concise and easy to understand." The Thai text in the image translates as follows: "\\Which account will the money be refunded to? \\The money will be refunded to the account used for the purchase.\\ What is the procedure if I want to change the product? \\ To exchange a product, first contact the seller or their representative to inform them of your intention to exchange. You will then receive a reference number. Afterward, reach out again to the seller or representative, provide the details of the product you wish to exchange, and wait to receive the new product based on the reference number." }
\caption{The model performs well in e-commerce Q\&A, providing concise and clear answers. }
\label{fig:efficient_and_concise_answers}
\end{figure}

We found that \llm performs effectively in the e-commerce QA context. Feedback from students at Chulalongkorn University described it as "concise and easy to understand." The Thai text in the Figure ~\ref{fig:efficient_and_concise_answers}  translates as follows: 

\begin{quote}
User: To which account will the refund be issued? \newline
Assistant: The refund will be issued to the account used for the purchase. \newline
User: What is the procedure if I want to exchange the product? \newline
Assistant:  To exchange a product, first contact the seller or their representative to inform them of your intention. You will then receive a reference number. After that, reach out again to the seller or representative, provide details of the product you wish to exchange, and await the new product based on the reference number.  \newline
\end{quote}

\begin{figure}[ht]
\centering
\includegraphics[width=0.8\linewidth]{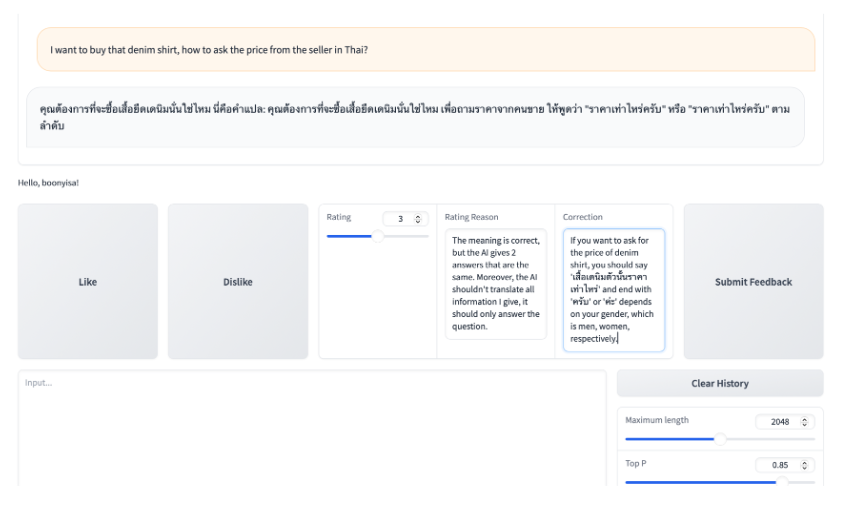}
% \caption{We observed that some open-source models, including ours, sometimes have difficulty distinguishing between male and female language in Thai. For example, gendered particles—such as the male-specific particle used at the end of sentences, and a different particle used in female speech—can be challenging for the model to differentiate.}
\caption{The model struggles with distinguishing gendered language in Thai, particularly with gendered particles.}

\label{fig:sexual_species_bad_case}
\end{figure}

However, as shown in Figure ~\ref{fig:sexual_species_bad_case}, the model struggled with gendered language in Thai, particularly with gendered particles. This issue was also observed in other state-of-the-art models, highlighting a common challenge in handling Thai gender distinctions.

\begin{figure}[ht]
\centering
\includegraphics[width=0.8\linewidth]{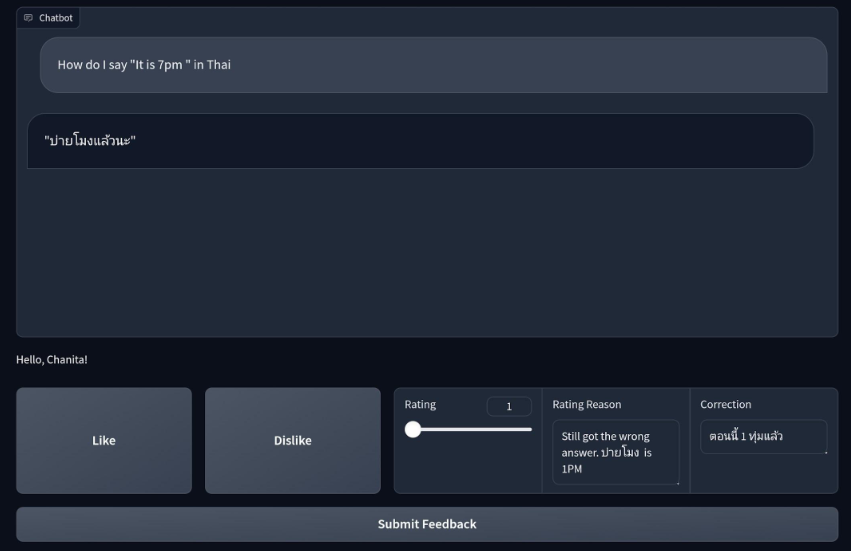}
% \caption{The LLM reveals that the unique ways of expressing time and numerical data in Thai present challenges in translation, often leading to inaccuracies and misrepresentation of content across languages. These nuances make it difficult for models to maintain precision in temporal and numerical data handling.}
\caption{The model faces challenges in handling time and numerical data in Thai, leading to translation errors.}
\label{fig:time_bad_case}
\end{figure}

Additionally, the LLM revealed that the unique ways of expressing time and numerical data in Thai pose challenges during translation, often resulting in inaccuracies and misrepresentation across languages. These subtleties make it difficult for models to handle temporal and numerical data with precision. Figure ~\ref{fig:time_bad_case} illustrates how the model misrepresented time and numerical data.

In summary, while our model performed well in e-commerce tasks, feedback highlighted its limitations in handling gender, politeness, and time-related expressions in Thai. These areas require further refinement.
% \FloatBarrier % Ensures that all figures are placed before proceeding

\subsection{Thai-Specific Evaluation Dataset: Xdata\_Thai}

In collaboration with Chulalongkorn University, we developed a Thai-specific evaluation dataset comprising 350 manually annotated samples. Created through ad hoc testing ~\ref{sec:evaluation_sights}, peer reviews, and consensus-based annotations, the dataset ensures cultural and linguistic accuracy. It addresses low-resource language challenges by focusing on Thai idioms, slang, and formal tone distinctions. The task distribution, shown in Figure~\ref{fig:evaluation_set_task_type_ratio}, highlights practical applications like e-commerce and idiomatic language, which are often overlooked in existing benchmarks.

To evaluate both pre-trained and instruction-tuned models, we formatted the tasks as continuation prompts in a 3-shots setting. The prompt format is detailed in Appendix ~\ref{sec:appendix_evaluation_set}. Table~\ref{tab:xdata_thai_v2} presents the evaluation results, demonstrating the effectiveness of our instruction-tuned model in handling low-resource scenarios.

% Detailed descriptions of the prompt format and evaluation procedure can be found in Appendix~\ref{sec:appendix}.

\subsubsection{Creation of Evaluation Dataset}
\label{sec:appendix_evaluation_set}
The development of the Thai-specific evaluation dataset was a meticulous, multi-step process aimed at ensuring cultural and linguistic precision. This dataset was designed to benchmark the performance of large language models (LLMs) in Thai, with a particular focus on cross-border e-commerce use cases. The dataset creation process consisted of the following phases:

\subsubsubsection{Question Collection Process}

\begin{figure}[ht]
\centering
\includegraphics[width=0.8\linewidth]{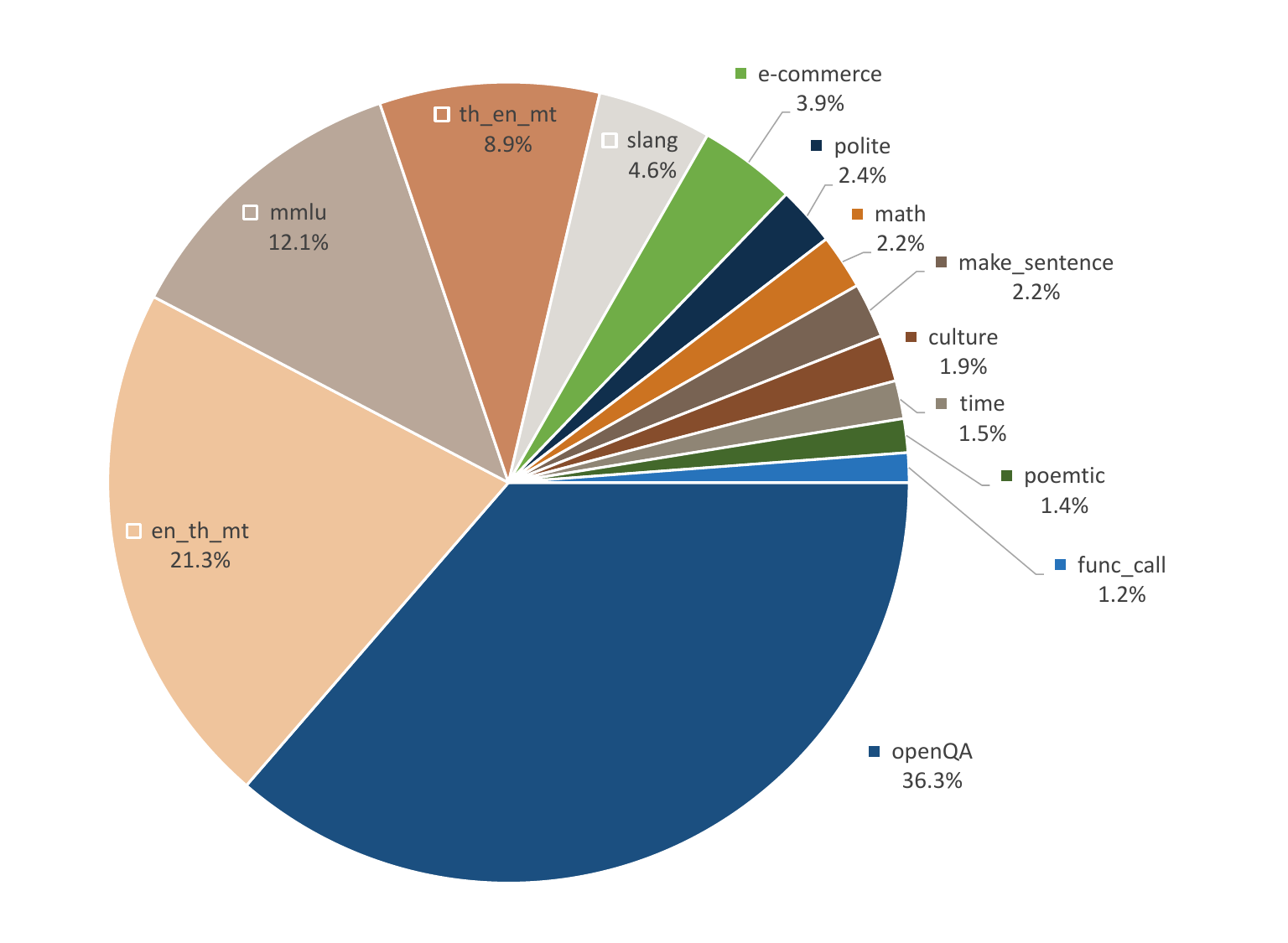}
\caption{The distribution of task types in our evaluation set. Task types with fewer than 5 occurrences were removed.}
\label{fig:evaluation_set_task_type_ratio}
\end{figure}

\begin{enumerate}
    \item \textbf{Ad Hoc Testing}: Thai university students were engaged to evaluate preliminary model outputs and suggest corrections. This process, described in Appendix~\ref{sec:appendix}, served as the foundation for question refinement.
    \item \textbf{Peer Review}: Native Thai speakers in three independent groups reviewed the initial corrections and model outputs. These reviewers provided ratings, feedback, and alternative suggestions to enhance both linguistic and contextual accuracy.
     \item \textbf{Final Selection}: The best answers were selected based on a thorough collaborative review process involving manual filtering and comprehensive quality assessment. Each question was then categorized by task type, as shown in Figure~\ref{fig:evaluation_set_task_type_ratio}. This dataset includes an extensive collection of Thai idioms and proverbs, addressing nuanced aspects of language comprehension often neglected in existing benchmarks. Moreover, a significant subset comprises e-commerce-related questions, emphasizing practical application in cross-border trade scenarios.
\end{enumerate}

\subsubsubsection{Design Principles}

To ensure the generation of high-quality questions, the following principles were adhered to \citep{SciQ}:
\begin{itemize}
    \item \textbf{No Yes/No Questions}: Questions were required to involve complex, meaningful tasks to challenge model reasoning.
    \item \textbf{Context Independence}: Questions were designed to be standalone, not requiring additional context for understanding.
    \item \textbf{Minimization of Ambiguity}: Careful attention was given to avoid vague or overly open-ended phrasing.
\end{itemize}
   
\subsubsubsection{Distractor Design Strategy}

High-quality distractors were critical for evaluating model robustness. The process of generating domain-relevant incorrect options followed these steps:

\begin{itemize}
    \item \textbf{Leveraging Original Outputs}: Model-generated responses and corrections rejected during peer review were used as primary sources for distractors. Although suboptimal, these responses captured realistic errors, including both inaccuracies from the model and imperfect reviewer suggestions.
    \item \textbf{Supplementing with Generated Errors}: For missing distractors, errors were generated using an instruct-tuned version of XmodelLM \citep{wang2024xmodellmtechnicalreport}. This model was trained on a dataset that excluded Thai-specific instructions, ensuring distractors were plausible yet domain-appropriate.
    \item \textbf{Cleaning and Replacement}: Repetitive, overly verbose, or low-quality responses were either cleaned or replaced. Replacements were drawn from actual  e-commerce QA examples or other contextually relevant incorrect answers.
    % Replacements were sourced from actual Thai e-commerce QA examples or other contextually appropriate incorrect answers, including GPT-translated outputs from comprehensive datasets and incorrect answers found in the evaluation set.
\end{itemize}

% For distractor selection, we prioritized using the original model outputs and corrections that were not adopted during the review process. These corrections, though suboptimal, included both model-generated responses and imperfect suggestions from reviewers. Missing distractors were supplemented with error examples generated by an instruct-tuned version of XmodelLM, which was trained on data excluding Thai-specific instructions, ensuring domain relevance. Repetitive or overly lengthy responses were cleaned or replaced with translations from actual Thai e-commerce QA examples or other contextually appropriate incorrect answers.

This rigorous process ensured that the dataset captures both linguistic subtleties and practical relevance, establishing it as a comprehensive benchmark for evaluating models in low-resource languages. To maximize its utility, we developed a structured evaluation framework encompassing task design, tailored evaluation metrics, and detailed data analysis, specifically suited for both pre-trained and instruction-tuned models.

\subsubsection{Evaluation Task Design}
\begin{figure}[ht]
\centering
\begin{minipage}{0.435\linewidth}
    \centering
    \includegraphics[width=\linewidth]{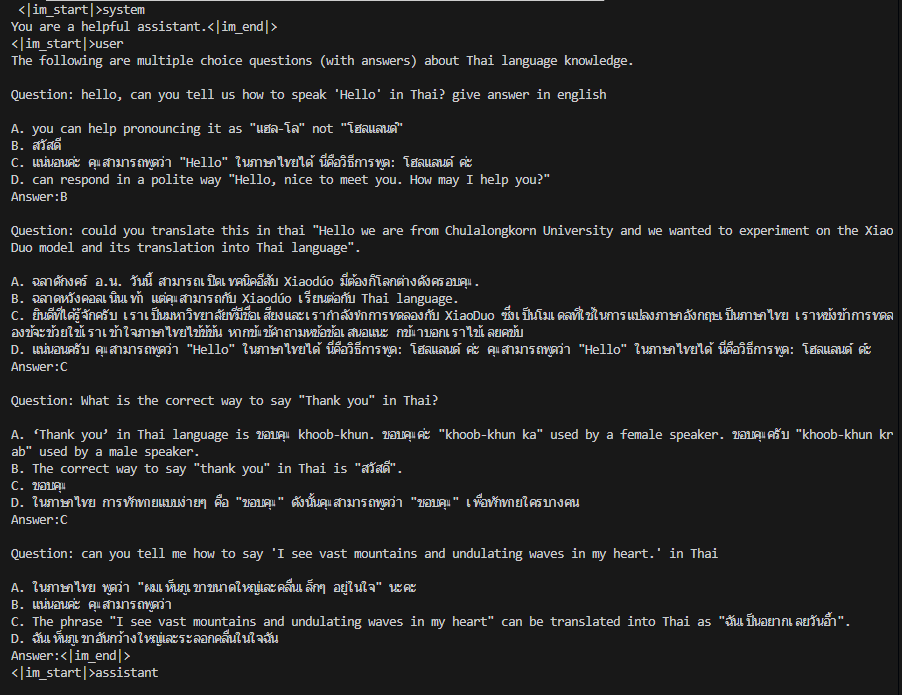}
    \caption{Example of a multiple-choice prompt incorporating the chat template.}
    \label{fig:example_prompt_with_chat_template}
\end{minipage}
\hfill
\begin{minipage}{0.49\linewidth}
    \centering
    \includegraphics[width=\linewidth]{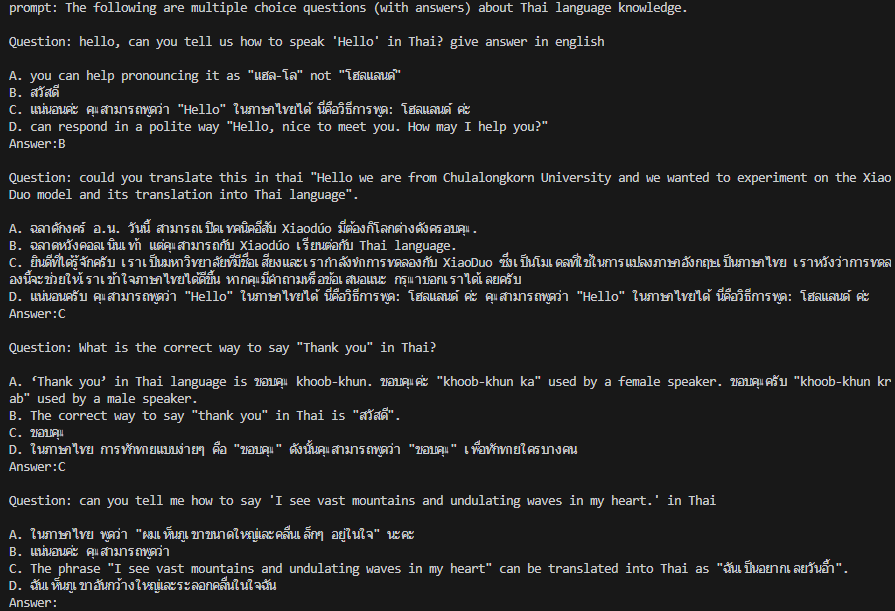}
    \caption{Example of a multiple-choice prompt without the chat template.}
    \label{fig:example_prompt}
\end{minipage}
\end{figure}

To assess model performance, we designed the evaluation tasks as ABCD multiple-choice continuation prompts. The task design and evaluation methodology are as follows:

\begin{itemize}
    \item \textbf{Randomized Options}: The answer options (A, B, C, D) are randomized for each question to mitigate positional bias.
    \item \textbf{Token-Based Matching}: The model generates a response continuation, and its predicted choice is determined from the first 10 tokens. This prediction is compared to the correct answer to calculate accuracy.
    \item \textbf{Few-Shot and and Chat-Adapted Evaluation}: The framework supports few-shot learning settings and optionally integrates chat templates to align more closely with the conversational nature of certain models.
\end{itemize}
The prompt format used for the evaluation is shown below:
\begin{center}
\begin{minipage}{0.6\linewidth} % 将宽度设置为页面的 80%
\begin{lstlisting}
The following are multiple choice questions (with answers) about Thai language knowledge.

Question: {{Question1}}  
A. {{A}}  
B. {{B}}  
C. {{C}}  
D. {{D}}  
Answer: {{right_answer}}  

Question: {{Question2}}  
A. {{A}}  
B. {{B}}  
C. {{C}}  
D. {{D}}  
Answer: {{right_answer}}  

Question: {{Question3}}  
A. {{A}}  
B. {{B}}  
C. {{C}}  
D. {{D}}  
Answer: {{right_answer}}  

Question: {{Question}}  
A. {{A}}  
B. {{B}}  
C. {{C}}  
D. {{D}}  
Answer:
\end{lstlisting}
\end{minipage}
\end{center}
Figures~\ref{fig:example_prompt_with_chat_template} and \ref{fig:example_prompt} provide visual examples of these prompts. They illustrate how answer options are shuffled and how the generated continuations are compared with the correct answers for evaluation.

% \begin{figure}[ht]
% \centering
% \includegraphics[width=\linewidth]{figures/example_prompt_with_chat_template.png}
% \caption{Example of a multiple-choice prompt using the chat template format.}
% \label{fig:example_prompt_with_chat_template}
% \end{figure}

% \begin{figure}[ht]
% \centering
% \includegraphics[width=\linewidth]{figures/example_prompt.png}
% \caption{Example of a multiple-choice prompt without the chat template.}
% \label{fig:example_prompt}
% \end{figure}

% \subsubsection{Data Analysis}
% To further evaluate the dataset's performance across models, we conducted extensive data analysis. Results highlight model consistency and discrimination capabilities. Figure~\ref{fig:model_analysis} presents the comparison of accuracy and variance across multiple models, emphasizing the dataset's utility in benchmarking low-resource language tasks.

In the table below, we compare the performance of PolyLM-1.7B and \llm-1B on the Xdata\_Thai evaluation dataset, using a 3-shot setting.

% \FloatBarrier % Ensures that all figures are placed before proceeding
\begin{table}[ht]
  \centering
  \setlength{\tabcolsep}{10pt}

  \label{tab:thai_language_tasks}
    \centering
    \begin{tabular}{lc}
      \toprule
      Model & \textbf{Xdata\_Thai (3-shots)} \\
      \midrule
      % llama3.2-1B-Instruct & 0.424 \\
      % llama3.2-3B-Instruct & 0.646 \\
      % Qwen2-1.5B-Instruct  & 0.605 \\
      PolyLM-1.7B          & 0.228 \\
       % \midrule
       \llm-1B                & 0.237\\
      % \llm-1B-Instruct      & 0.307 \\
      \bottomrule
    \end{tabular}
    \caption{Performance comparison on Xdata\_Thai}
    \label{tab:xdata_thai_v2}
\end{table}
\subsection{Detailed Evaluation Results}
\label{sec:appendix-2}
This section presents a comprehensive evaluation of the model's performance on multilingual tasks, focusing on Thai, Arabic, French, and Chinese. The evaluation results are reported using standard accuracy (acc) metrics, offering a clear comparison of the model's capabilities across these diverse languages.

\begin{table}[ht]
  \centering
  \setlength{\tabcolsep}{10pt}
  \caption{Performance on multilingual tasks (Thai, Arabic, French, Chinese).}
  \label{tab:multilingual_tasks}
      \begin{subtable}[t]{\textwidth}
    \centering
    \begin{tabular}{lcc}
      \toprule
      Model & \textbf{belebele\_tha\_Thai} & \textbf{xcopa\_th} \\
      \midrule
      PolyLM-1.7B  & 0.2267 & 0.56 \\
      PolyLM-13B   & 0.2367 & 0.586 \\
      \midrule
      \llm-1B & 0.2756 & 0.59 \\
      \bottomrule
    \end{tabular}
    \caption{Performance on Thai language tasks.}
    \label{tab:thai}
  \end{subtable}
  
  \vspace{10pt} % Add spacing between subtables

  \begin{subtable}[t]{\textwidth}
    \centering
    \begin{tabular}{lcccc}
      \toprule
      Model & \textbf{arc\_ar} & \textbf{hellaswag\_ar} & \textbf{m\_mmlu\_ar} & \textbf{piqa\_ar} \\
      \midrule
      PolyLM-1.7B  & 0.2173 & 0.2818 & 0.2288 & 0.5381 \\
      PolyLM-13B   & 0.2284 & 0.3296 & 0.2434 & 0.5653 \\
      \midrule
      \llm-1B & 0.2344 & 0.3279 & 0.2454 & 0.5789 \\
      \bottomrule
    \end{tabular}
    \caption{Performance on Arabic language tasks.}
    \label{tab:arabic}
  \end{subtable}

  \vspace{10pt} % Add spacing between subtables
\begin{subtable}[t]{\textwidth}
    \centering
    \begin{tabular}{lcccc}
    \toprule
    Model & \textbf{hellaswag\_fr} & \textbf{m\_mmlu\_fr} & \textbf{paws\_fr} & \textbf{piqa\_fr} \\
    \midrule
    PolyLM-1.7B  & 0.3085 & 0.2458 & 0.548 & 0.5381 \\
    PolyLM-13B   & 0.4064 & 0.2602 & 0.539 & 0.5653 \\
    \midrule
    \llm-1B & 0.37 & 0.2525 & 0.5325 & 0.5789 \\
    \bottomrule
  \end{tabular}
  \caption{Performance on French language tasks.}
  \label{tab: french}
 \end{subtable}
    \vspace{10pt} % Add spacing between subtables
  \begin{subtable}[t]{\textwidth}
    \centering
    \begin{tabular}{lcccc}
      \toprule
      Model & \textbf{arc\_zh} & \textbf{xcopa\_zh}  \\
      \midrule
      PolyLM-1.7B  & 0.1957 & 0.5381  \\
      PolyLM-13B   & 0.2803 & 0.5653 \\
      \midrule
      \llm-1B & 0.259 & 0.5789  \\
      \bottomrule
    \end{tabular}
    \caption{Performance on Chinese language tasks.}
    \label{tab:chinese}
  \end{subtable}

\end{table}

\end{document}